\documentclass[10pt,twocolumn,twoside,journal]{IEEEtran}

\usepackage[noadjust]{cite}

\usepackage[noadjust]{cite}

\ifCLASSINFOpdf
\usepackage[pdftex]{graphicx}
\DeclareGraphicsExtensions{.pdf,.jpeg,.png,.tif,.tiff}
\else

\fi

\usepackage[cmex10]{amsmath}

\ifCLASSOPTIONcompsoc
\usepackage[tight,normalsize,sf,SF]{subfigure}
\else
\usepackage[tight,footnotesize]{subfigure}
\fi
\usepackage{subfigure}
\usepackage[latin1]{inputenc}
\usepackage{amsthm, amsfonts, amssymb, mathrsfs, amsmath}
\usepackage{amsmath}
{
	\theoremstyle{plain}
	
}
\usepackage{nicefrac}
\usepackage{mathtools}
\usepackage{multirow}
\usepackage[ruled,linesnumbered]{algorithm2e}
\let\oldnl\nl
\newcommand{\nonl}{\renewcommand{\nl}{\let\nl\oldnl}}
\usepackage[mathcal]{euscript}
\usepackage[draft]{fixme}
\usepackage{enumerate}
\usepackage{dsfont}
\usepackage{tikz}
\allowdisplaybreaks[1]

\newcommand{\bM}{{\boldsymbol{M}}}
\newcommand{\bL}{{\boldsymbol{L}}}
\newcommand{\bS}{{\boldsymbol{S}}}
\newcommand{\bZ}{{\boldsymbol{Z}}}
\newcommand{\bB}{{\boldsymbol{B}}}
\newcommand{\bT}{{\boldsymbol{\Theta}}}

\newcommand{\bU}{{\boldsymbol{U}}}
\newcommand{\bV}{{\boldsymbol{V}}}
\newcommand{\bSi}{{\boldsymbol{\Sigma}}}

\newcommand{\bX}{{\boldsymbol{X}}}
\newcommand{\bx}{{\boldsymbol{x}}}

\newcommand{\bv}{{\boldsymbol{v}}}
\newcommand{\bz}{{\boldsymbol{z}}}
\newcommand{\by}{\boldsymbol{y}}

\newcommand{\bdel}{\boldsymbol{\delta}}
\newcommand{\be}{\boldsymbol{e}}

\newcommand{\bhx}{{\boldsymbol{\widehat{x}}}}
\newcommand{\bhv}{{\boldsymbol{\widehat{v}}}}

\makeatletter
\renewcommand{\@algocf@capt@plain}{above}
\makeatother

\DeclareMathOperator*{\argmin}{arg\,min}

\usepackage{amsthm}
\usepackage{float}

\begin{document}
\title{Compressive Online Robust Principal Component Analysis with Optical Flow for Video Foreground-Background Separation}


\author{Srivatsa Prativadibhayankaram, Huynh Van Luong*, Thanh-Ha Le,~and~Andr\'{e} Kaup

	\thanks{S. Prativadibhayankaram, H. V. Luong, and A. Kaup are with the Chair of Multimedia Communications and Signal Processing, Friedrich-Alexander-Universit\"{a}t Erlangen-N\"{u}rnberg, 91058 Erlangen, Germany (e-mail: srivatsa.pv@live.com, huynh.luong@fau.de, and andre.kaup@fau.de).}
	\thanks{Thanh-Ha Le is with the Human Machine Interaction Lab, University of Engineering and Technology, Vietnam National University, Hanoi, Vietnam (e-mail: ltha@vnu.edu.vn).}
	\thanks{ *Corresponding author. Tel: +49 9131 85 27664. Fax: +49 9131 85 28849. E-mail address: huynh.luong@fau.de (H. V. Luong).}
}




\maketitle
\begin{abstract}
	In the context of online Robust Principle Component Analysis (RPCA) for the video foreground-background separation, we propose a compressive online RPCA with optical flow that separates recursively a sequence of frames into sparse (foreground) and low-rank (background) components. Our method considers a small set of measurements taken per data vector (frame), which is different from conventional batch RPCA, processing all the data directly. The proposed method also incorporates multiple prior information, namely previous foreground and background frames, to improve the separation and then updates the prior information for the next frame. Moreover, the foreground prior frames are improved by estimating motions between the previous foreground frames using optical flow and compensating the motions to achieve higher quality foreground prior. The proposed method is applied to online video foreground and background separation from compressive measurements. The visual and quantitative results show that our method outperforms the existing methods.
\end{abstract}
\section{Introduction}\label{intro}

The background and foreground separation of a video sequence is of great importance in a number of computer vision applications, e.g., visual surveillance and object detection. These separations make the video analysis more efficient and regions of interest extracted can be used as a preprocessing step for further identification and classification. In video separation, a video sequence is separated into the slowly-changing background (modeled by $\bL$ as a low-rank component) and the foreground (modeled by $\bS$ as a sparse component). Robust Principle Component Analysis (RPCA) \cite{CandesRPCA,JWright09} was shown to be a robust method for separating the low-rank and sparse compenents. RPCA decomposes a data matrix $\bM $ into the sum of unknown sparse $ \bS $ and low-rank $\bL$ by solving the Principal Component Pursuit (PCP) \cite{CandesRPCA} problem:
\begin{equation}\label{PCP}
\min_{\bL,\bS} \|\bL\|_{*}+\lambda\|\bS\|_{1} \mathrm{~subject~to~}\bM = \bL + \bS,
\vspace{-0.4pt}
\end{equation}
where $\|\cdot\|_{*}$ is the matrix nuclear norm (sum of singular values) and $\|\cdot\|_1$ is the $\ell_1$-norm. RPCA has found many applications in computer vision, web data analysis,
and recommender systems. However, batch RPCA processes all data samples, e.g., all frames in a video, which involves high computational and memory requirements. 

Moreover, with inherent characteristics of video, correlations among consecutive frames can be taken into account to improve the separation. The correlations can be obtained in the form of motions that present the information changes from one frame to the others. Detecting motion is an integral part of the human visual system. One of the dominant techniques for estimating motion in computer vision is optical flow by variational methods \cite{Horn1981,Bruhn2005,Baker2011}. The optical flow estimates the motion vectors of all pixels in a given frame due to the relative motions between frames. In particular, the motion vectors at each pixel can be estimated by minimizing a gradient-based matching of pixel gray value that is combined with a smoothness criteria \cite{Horn1981}. Thereafter, the computed motion vectors in the horizontal and vertical directions \cite{bruhn2005lucas} are used to compensate and predict information in the next frame. For producing highly accurate motions and correct large displacement correspondences, a large displacement optical flow \cite{brox2011} combines a coarse-to-fine optimization with descriptor matching. Therefore, the large displacement optical flow \cite{brox2011} can be exploited in the video separation to estimate the motions from previously separated frames to support the current frame separation.

%

In order to deal with the video separation in an online manner, we consider an online RPCA algorithm that recursively processes a sequence of frames (a.k.a., the column-vectors in $\bM$) per time instance. Additionally, we aim at recovering the foreground and background from a small set of measurements rather than a full frame data, leveraging information from a set of previously separated frames. In particular, at time instance $t$, we wish to separate $\bM_{t} $ into $\bS_{t}\hspace{-2pt}=\hspace{-2pt}[\textit{\textbf{x}}_{1}~ \bx_{2}~ ...~ \bx_{t}]$ and $\bL_{t}\hspace{-2pt}=\hspace{-2pt}[\bv_{1}~ \bv_{2}~ ...~ \bv_{t}]$, where $[\cdot]$ denotes a matrix and $\bx_{t}, \bv_{t}\in \mathbb{R}^{n}$ are column-vectors in $\bS_{t}$ and $\bL_{t}$, respectively. We assume that $\bS_{t-1}\hspace{-2pt}=\hspace{-2pt}[\textit{\textbf{x}}_{1}~ \bx_{2}~ ...~ \bx_{t-1}]$ and $\bL_{t-1}\hspace{-2pt}=\hspace{-2pt}[\bv_{1}~ \bv_{2}~ ...~ \bv_{t-1}]$ have been recovered at time instance $t\hspace{-0,5pt}-\hspace{-0,5pt}1$ and that at time instance $t$ we have access to compressive measurements of the full frame, a.k.a., vector $\bx_{t}+\bv_{t}$, that is, we observe $ \by_{t} = \mathbf{{\Phi}}(\bx_{t} + \bv_{t})$, where $\mathbf{{\Phi}}\in \mathbb{R}^{m\times n} (m < n) $ is a random projection. 
The recovery problem at time instance $t$ is thus written \cite{LuongARXIV17} as
\begin{equation}\label{onlinePCP}
\min_{\bx_{t},\bv_{t}} \|[\bL_{t-1}~\bv_{t}]\|_{*}\hspace{-2pt}+\hspace{-1pt}\lambda\|\bx_{t}\|_{1} \mathrm{~subject~to~}\textbf{\textit{y}}_{t}\hspace{-2pt}=\hspace{-2pt}\mathbf{\Phi}(\bx_{t}+\bv_{t}),
\end{equation}
where $\bL_{t-1}$, $\bS_{t-1}$, and $\mathbf{\Phi}$ are given.

There are several works on the separation problems \cite{Rodriguez16,Rodriguez16,JHe12,JXu13,Feng13,Mansour15} by advancing RPCA \cite{CandesRPCA}. Incremental PCP \cite{Rodriguez16} processes each column-vector in $\bM$ at a time. However, assuming access to the complete data (e.g., full frames) rather than compressive data. On the other hand, Compressive PCP \cite{JWright13} is a counterpart of batch RPCA that operates on compressive measurements. Some studies in \cite{JHe12,JXu13,Feng13,Mansour15}
addressed the problem of online estimation of low-dimensional subspaces from randomly subsampled data for modeling the background. The work in \cite{GuoQV14} proposed an algorithm to recover the sparse component $\bx_t$ in~\eqref{onlinePCP}, however, the low-rank component $\bv_{t}$ in~\eqref{onlinePCP} was not recovered per time instance from a small number of measurements. 
The alternative method in \cite{MotaTSP16}, \cite{WarnellTIP15} estimates the number of compressive measurements required to recover foreground $\bx_t$ per time instance via assuming the background $\bv_t$ not-varying. This assumption is invalid in realistic scenarios due to illumination variations or moving backgrounds.

The problem of separating a sequence of time-varying frames using prior information brings significant improvements in the context of online RPCA~\cite{GuoQV14,QiuVLH14,Zhan14ISIT}. Several studies on recursive recovery from low-dimensional measurements have been proposed to leverage prior information~\cite{MotaTSP16,GuoQV14,QiuVLH14,VaswaniZ16}. The study in~\cite{VaswaniZ16} provided a comprehensive overview of the domain, reviewing a class of recursive algorithms. 
The studies in~\cite{GuoQV14,QiuVLH14} used modified-CS~\cite{NVaswaniTSP10} to leverage prior knowledge under the condition of slowly varying support and signal values. However, this method as well as the methods in~\cite{JHe12,JXu13,Mansour15} do not explore the correlations between the current frame and multiple previously separated frames. Our latest work in \cite{LuongARXIV17} leverages correlations across the previously separated foreground frames. However, displacements between the previous foreground frames and the current frame are not taken into account. These displacements can incur the degradation of the separation performance. 

\textbf{Contribution}. We propose a \textit{compressive online robust PCA with optical flow} (CORPCA-OF) method, which is based on our previous work in \cite{LuongARXIV17}, to leverage information from previously separated foreground frames via optical flow \cite{brox2011}. The novelty of CORPCA-OF over CORPCA \cite{LuongARXIV17} is that the optical flow is used to estimate and compensate motions between the foreground frames to generate new prior foreground frames. These new prior frames have high correlation with the current frame and thus improve the separation. We also exploit the slowly-changing characteristics of backgrounds known as low-rank components via an incremental $\mathrm{SVD}$ \cite{Brand02} method. The compressive separation problem in~\eqref{onlinePCP} is solved in an online manner by minimizing not only an $n$-$\ell_{1}$-norm cost function \cite{LuongICIP16} for the sparse foreground but also the rank of a matrix for the low-rank backgrounds. Thereafter, the new separated foreground and background frames are used to update the prior knowledge for the next processing instance. 

The rest of this paper is organized as follows. We summarize the CORPCA algorithm \cite{LuongARXIV17}, on which our proposed method is to be built, and state our problem in Sec. \ref{problem}. The proposed method is fully described in Sec. \ref{corpca-of}. We test our proposed method for an online compressive video separation application on real video sequences and evaluate both visual and quantitative results in Sec. \ref{Experiment}. 

\section{Video foreground-background separation using Compressive Online Robust PCA with Optical Flow}\label{VideoFGBGSep}
In this section, we firstly review the CORPCA algorithm \cite{LuongARXIV17} for online compressive video separation and state our problem. Thereafter, we propose the CORPCA-OF method, which is summarized in the CORPCA-OF algorithm. 
\subsection{Compressive Online Robust PCA (CORPCA) for Video Separation}\label{problem}
\begin{figure*}[tp!]
	\centering
	\includegraphics[width = 0.73\textwidth]{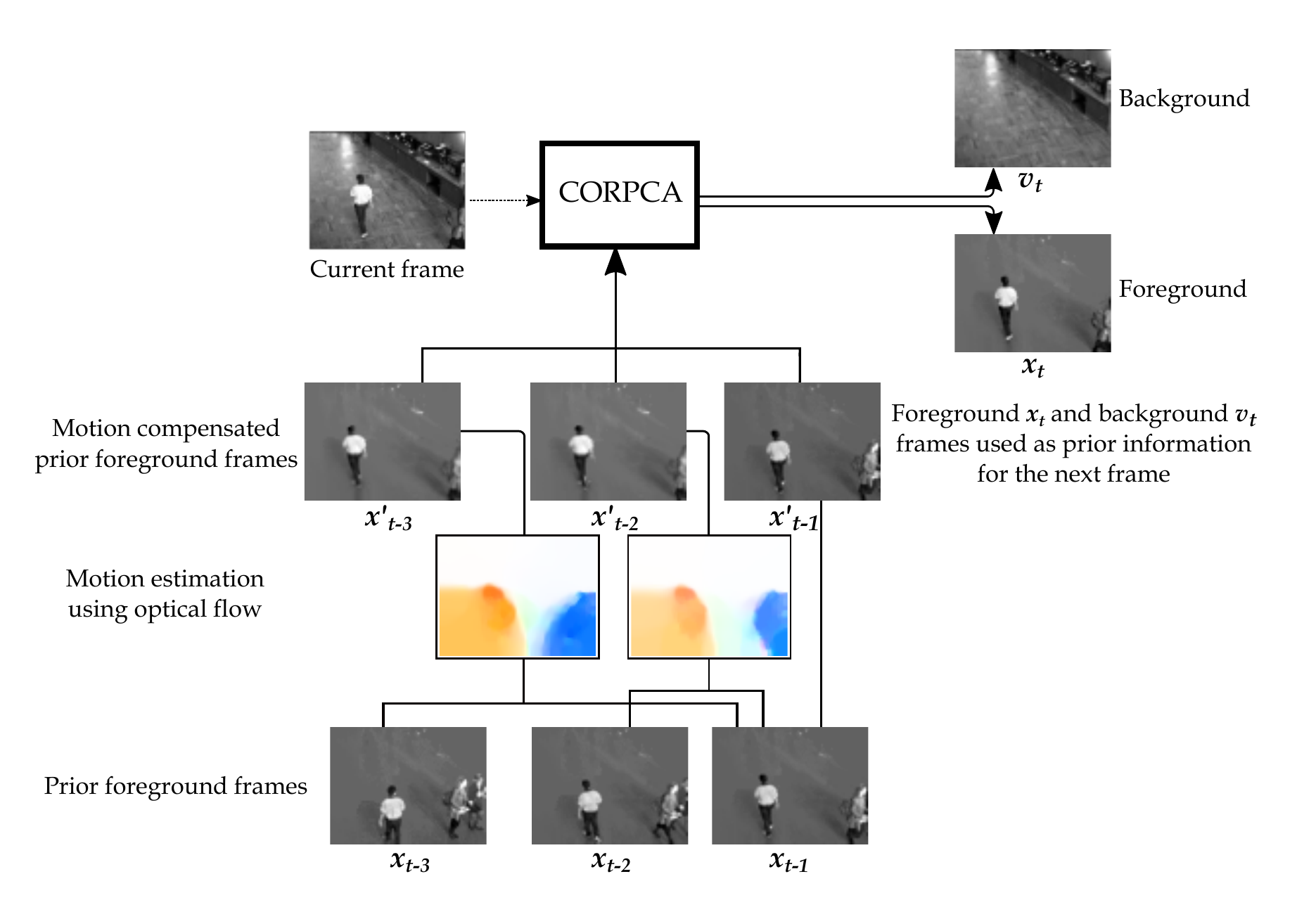}
	\caption{CORPCA-OF work flow.}
	\label{CORPCA-OF-blockDiagram}
\end{figure*}
The CORPCA algorithm \cite{LuongARXIV17} is proposed for video separation that is based on the RAMSIA algorithm \cite{LuongICIP16} solving an $n$-$\ell_1$ minimization problem with adaptive weights to recover a sparse signal $\bx$ from low-dimensional random measurements $\by=\mathbf{\Phi} \bx$ with the aid of multiple prior information $\bz_j$, $j\in\{0,1,\dots,J\}$, with $\bz_{0}=\mathbf{0}$. 
The objective function of RAMSIA \cite{LuongICIP16} is given by
\vspace{-0.9pt}
\vspace{-0.10pt}
\vspace{-0.8pt}
\begin{equation}\label{n-l1minimizationGlobal}
\min_{\bx}\hspace{-0pt}\Big\{\hspace{-0pt}H(\bx)\hspace{-2pt}=\hspace{-2pt}\frac{1}{2}\|\mathbf{\Phi}\bx-\by\|^{2}_{2} + \lambda \hspace{-2pt}\sum\limits_{j=0}^{J}\hspace{-2pt}\beta_{j}\|\mathbf{W}_{j}(\bx-\bz_{j})\|_{1}\Big\},
\vspace{-0.8pt}
\vspace{-0.6pt}
\end{equation}
where $\lambda>0$ and $\beta_{j}\hspace{-2pt}>\hspace{-2pt}0$ are weights across the prior information, and $\mathbf{W}_{j}$ is a diagonal matrix with weights for each element in the prior information signal $\bz_{j}$; namely, $\mathbf{W}_{j}\hspace{-2pt}=\hspace{-2pt}\mathrm{diag}(w_{j1},w_{j2},...,w_{jn})$ with $w_{ji}\hspace{-2pt}>\hspace{-2pt}0$ being the weight for the $i$-th element in the $\bz_{j}$ vector.

The CORPCA algorithm processes one data vector per time instance by leveraging prior information for both its sparse and low-rank components. At time instance $t$, we observe $\by_{t}=\mathbf{\Phi}(\bx_{t}+\bv_{t})$ with $\by_{t}\in\mathbb{R}^{ m}$.
Let $\bZ_{t-1}:=\{\bz_{1},...,\bz_{J}\}$, a set of $\bz_{j}\in \mathbb{R}^{n}$, and $\bB_{t-1}\in \mathbb{R}^{n\times d}$ denote prior information for $\bx_{t}$ and $\bv_{t}$, respectively. The prior information $\bZ_{t-1}$ and $\bB_{t-1}$ are formed by using the already reconstructed set of vectors $\{ \hat\bx_{1}, ..., \hat\bx_{t-1}\} $ and $\{\hat\bv_{1}, ..., \hat\bv_{t-1}\}$.

The objective function of CORPCA is to solve Problem \eqref{onlinePCP} and formulated by
\vspace{-0.7pt}
\begin{align}\label{CORPCAminimization}
\hspace{-2pt}\min_{\bx_{t},\bv_{t}}\hspace{-0pt}&\Big\{\hspace{-0pt}H(\bx_{t},\bv_{t}|\by_{t},\bZ_{t-1},\bB_{t-1})\hspace{-2pt}=\hspace{-2pt}\frac{1}{2}\|\mathbf{\Phi}(\bx_{t}+\bv_{t})-\by_{t}\|^{2}_{2} \nonumber \\
\vspace{-0.15pt}
&+\lambda \mu\hspace{-2pt}\sum\limits_{j=0}^{J}\hspace{-2pt}\beta_{j}\|\mathbf{W}_{j}(\bx_{t}-\bz_{j})\|_{1}+\mu\Big\|[\bB_{t-1}~ \bv_{t}]\Big\|_{*}\Big\},
\vspace{-0.8pt}
\vspace{-0.23pt}
\end{align}
where $\mu>0$. It can be seen that when $\bv_{t}$ is static (not changing), Problem \eqref{CORPCAminimization} would become Problem \eqref{n-l1minimizationGlobal}. Furthermore, when $\bx_{t}$ and $\bv_{t}$ are batch variables 
and we do not take the prior information, $\bZ_{t-1}$ and $\bB_{t-1}$, and the projection $\mathbf{\Phi}$ into account, Problem \eqref{CORPCAminimization} becomes Problem \eqref{PCP}.

The CORPCA algorithm\footnote{The code of the CORPCA algorithm, the test sequences, and the corresponding outcomes are available at https://github.com/huynhlvd/corpca} solves Problem \eqref{CORPCAminimization} given that $\bZ_{t-1}$ and $\bB_{t-1}$ are known (they are obtained from the time instance or recursion). Thereafter, we update $\bZ_{t}$ and $\bB_{t}$, which are used in the following time instance. 

Let us denote $f(\bv_{t},\bx_{t})= (1/2)\|\mathbf{\Phi}(\bx_{t}+\bv_{t})-\by_{t} \|^{2}_{2}$,~$g(\bx_{t})=\lambda \hspace{-2pt}\sum_{j=0}^{J}\hspace{-2pt}\beta_{j}\|\mathbf{W}_{j}(\bx_{t}-\bz_{j})\|_{1}$, and $\textsl{h}(\bv_{t})=\|[\bB_{t-1}~ \bv_{t}]\|_{*}$. \hspace{1pt}As shown in the COPRCA algorithm \cite{LuongARXIV17}, $\bx_{t}^{(k+1)}$ and $\bv_{t}^{(k+1)}$ are iteratively computed at iteration $k+1$ via the soft thresholding operator \cite{Beck09} for $\bx_{t}$ and the single value thresholding operator \cite{Cai10} for $\bv_{t}$:
\vspace{-0.4pt}
\begin{equation}
\label{vtProximal}
\hspace{-0pt}\bv_{t}^{(\hspace{-1pt}k+1\hspace{-1pt})}\hspace{-2pt}=\hspace{-1pt}\argmin_{\bv_{t}}\hspace{-0pt}\Big\{\hspace{-0pt}\mu\textsl{h}(\bv_{t})\hspace{-1pt}+\hspace{-1pt}\Big\|\bv_{t}\hspace{-1pt}-\hspace{-1pt}\Big(\bv_{t}^{(\hspace{-1pt}k\hspace{-1pt})}\hspace{-2pt}-\hspace{-1pt}\frac{1}{2}\nabla_{\hspace{-2pt}\bv_{t}\hspace{-1pt}}f(\hspace{-1pt}\bv_{t}^{(\hspace{-1pt}k\hspace{-1pt})}\hspace{-1pt},\bx_{t}^{(\hspace{-1pt}k\hspace{-1pt})}\hspace{-1pt})\hspace{-1pt}\Big)\Big\|_{2}^{2}\hspace{-1pt}\hspace{-0pt}\Big\},\\\hspace{-7pt}
\vspace{-0.5pt}
\end{equation}
\vspace{-0.3pt}
\begin{equation}
\label{xtProximal}
\hspace{-0pt}\bx_{t}^{(\hspace{-1pt}k\hspace{-1pt}+1)}\hspace{-2pt}=\hspace{-1pt}\argmin_{\bx_{t}}\hspace{-0pt}\Big\{\hspace{-0pt}\mu g(\bx_{t})\hspace{-1pt}+\hspace{-1pt}\Big\|\bx_{t}\hspace{-1pt}-\hspace{-1pt}\Big(\bx_{t}^{(\hspace{-1pt}k\hspace{-1pt})}\hspace{-2pt}-\hspace{-1pt}\frac{1}{2}\nabla_{\hspace{-2pt}\bx_{t}} \hspace{-1pt}f(\hspace{-1pt}\bv_{t}^{(\hspace{-1pt}k\hspace{-1pt})}\hspace{-1pt},\bx_{t}^{(\hspace{-1pt}k\hspace{-1pt})}\hspace{-1pt})\hspace{-1pt}\Big)\Big\|_{2}^{2}\hspace{-1pt}\hspace{-0pt}\Big\}.\hspace{-8pt}
\vspace{-0.2pt}
\end{equation}

\textbf{Problem statement.} Using the prior information as in CORPCA \cite{LuongARXIV17} has provided the significant improvement of the current frame separation. However, there can be displacements between the consecutive frames deteriorating the separation performance. Fig. \ref{CORPCA-OF-blockDiagram} illustrates an example of three previous foreground frames, $\bx_{t-3},\bx_{t-2}$ and $\bx_{t-1}$. We can use them directly as prior information to recover foreground $\bx_{t}$ and background $\bv_{t}$ as done in CORPCA \cite{LuongARXIV17} due to the natural correlations between $\bx_{t}$ and $\bx_{t-3},\bx_{t-2},\bx_{t-1}$. In the last line of three prior foreground frames in Fig. \ref{CORPCA-OF-blockDiagram}, it can be seen that motions exist among them and the current frame $\bx_t$. By carrying out motion estimation using optical flow \cite{brox2011}, we can obtain the motions between the previous foreground frames as in Fig. \ref{CORPCA-OF-blockDiagram}, which are presented using color code for visualizing the motion flow fields \cite{brox2011}. These motions can be used to compensate and generate better quality prior frames (see compensated $\bx'_{t-3},\bx'_{t-2}$ compared with $\bx_{t-3},\bx_{t-2},\bx_{t-1}$), being more correlated to $\bx_{t}$. \hspace{1pt}In this work, we propose an algorithm - CORPCA with Optical Flow (CORPCA-OF), whose work flow is built as in Fig. \ref{CORPCA-OF-blockDiagram} by using optical flow \cite{brox2011} to improve prior foreground frames.

\subsection{The proposed COPRCA-OF Method}
\label{corpca-of}
\begin{figure}[tp!]
	\centering
	\setlength{\tabcolsep}{1pt}
	\renewcommand{\arraystretch}{0.1}
	\includegraphics[width=0.35\textwidth]{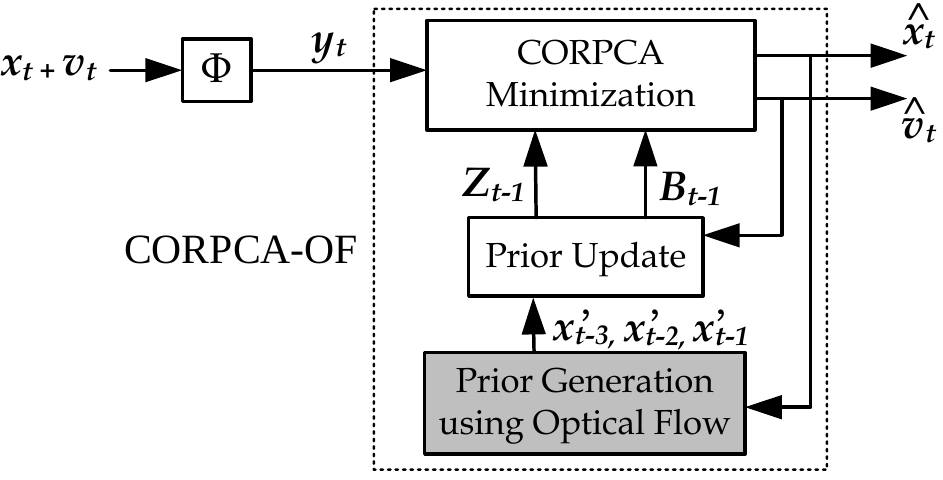}
	\caption{Compressive separation model using CORPCA-OF.}\label{corpcaModel}
	\vspace{-0.5pt}
\end{figure}

\textbf{Compressive Separation Model with CORPCA-OF}. Fig. \ref{corpcaModel} depicts a compressive separation model using the proposed CORPCA-OF method. Considering a time instance $t$, the inputs consist of compressive measurements $\by_{t}=\mathbf{\Phi}(\bx_{t}+\bv_{t})$ and prior information from time instance $t-1$, $\bZ_{t-1}$ and $\bB_{t-1}$. The model outputs foreground and background information $\bx_{t}$ and $\bv_{t}$ by solving the CORPCA minimization problem in \eqref{CORPCAminimization}. Finally, the outputs $\bx_{t}$ and $\bv_{t}$ are used to generate better prior foreground information via a prior generation using optical flow and update $\bZ_{t-1}$ and $\bB_{t-1}$ for the next instance via a prior update. The novel block of COPRCA-OF compared with CORPCA \cite{LuongARXIV17} is the Prior Generation using Optical Flow, where prior foreground information is improved by exploiting the large displacement optical flow \cite{brox2011}. The CORPCA-OF method is further described in Algorithm \ref{CORPCA-OFAlg}. 

\textbf{Prior Generation using Optical Flow}. CORPCA-OF aims at improving the foreground prior frames via optical flow. In Algorithm \ref{CORPCA-OFAlg}, the prior frames are initialized by $\bx_{t-1}$, $\bx_{t-2}$ and $\bx_{t-3}$. Optical flow is used to compute the motions between frames $\bx_{t-1}$ and $\bx_{t-3}$ (also $\bx_{t-1}$ and $\bx_{t-2}$) to obtain optical flow vectors for these two frames. This can be seen in the CORPCA-OF work flow diagram in Fig. \ref{CORPCA-OF-blockDiagram} as the optical flow fields represented in color code. The function $f_{ME}(\cdot)$ in Lines \ref{motion estimation 1} and \ref{motion estimation 2} [see Algorithm \ref{CORPCA-OFAlg}] computes the motions between prior foreground frames. This is based on the large displacement optical flow, as formulated in \cite{brox2011} and involves computing the optical flow vectors containing horizontal ($\mathsf{x}$) and vertical ($\mathsf{y}$) components, denoted by $\mathsf{\textbf{v}}_{1\mathsf{x}},\mathsf{\textbf{v}}_{2\mathsf{x}}$ and $\mathsf{\textbf{v}}_{1\mathsf{y}},\mathsf{\textbf{v}}_{2\mathsf{y}}$ $\in \mathbb{R}^{n}$, respectively. The estimated motions in the form of optical flow vectors, $(\mathsf{\textbf{v}}_{1\mathsf{x}},\mathsf{\textbf{v}}_{1\mathsf{y}})$ and $(\mathsf{\textbf{v}}_{2\mathsf{x}},\mathsf{\textbf{v}}_{2\mathsf{y}})$, are then used to predict the next frames by compensating for the forward motions on $\bx_{t-1}$. We generate the prior frames, $\bx'_{t-2}$ and $\bx'_{t-3}$, using motion compensation indicated by the function $f_{MC}(\cdot)$ as shown in Algorithm \ref{CORPCA-OFAlg} in Lines \ref{motion compensation 1} and \ref{motion compensation 2}.
 
Considering a point $i$ in the given frame, the horizontal and vertical components $\mathsf{v}_{1{\mathsf{x}i}}$ and $\mathsf{v}_{1\mathsf{y}i}$ of corresponding $\mathsf{\textbf{v}}_{1\mathsf{x}}$ and $\mathsf{\textbf{v}}_{1\mathsf{y}}$ are obtained, as mentioned in \cite{szeliski2010computer} by solving :
\begin{equation}\label{OpticalFlowEquation}
I_{1\mathsf{x}} \cdot \mathsf{v}_{1\mathsf{x}i} + I_{1\mathsf{y}} \cdot \mathsf{v}_{1\mathsf{y}i} + I_{1t} =0,
\end{equation}
where $I_{1\mathsf{x}} = {\partial I_{1}}/{\partial \mathsf{x}}$ and $I_{1\mathsf{y}} = {\partial I_{1}}/{\partial \mathsf{y}}$ are the intensity changes in the horizontal ($\mathsf{x}$) and vertical ($\mathsf{y}$) directions, respectively, constituting the spatial gradients of the intensity level $I_{1}$; $I_{1t} = {\partial I_{1}}/{\partial t}$ is the time gradient, which is a measure of temporal change in the intensity level at point $i$. There are various methods \cite{Horn1981,brox2011,Bruhn2005,Baker2011} to determine $\mathsf{v}_{1\mathsf{x}i}$ and $\mathsf{v}_{1\mathsf{y}i}$. Our solution is based on large displacement optical flow \cite{brox2011}, that is a combination of global and local approaches to estimate all kinds of motion. It involves optimization and minimization of error by using descriptor matching and continuation method, which utilizes feature matching along with conventional optical flow estimation to obtain the flow field. We combine the optical flow components of each point $i$ in the image into two vectors $(\mathsf{\textbf{v}}_{1\mathsf{x}},\mathsf{\textbf{v}}_{1\mathsf{y}})$, i.e., the horizontal and the vertical components of the optical flow vector. Similarly we obtain $(\mathsf{\textbf{v}}_{2\mathsf{x}},\mathsf{\textbf{v}}_{2\mathsf{y}})$.

The estimated motions in the form of optical flow vectors are used along with the frame $\bx_{t-1}$ to produce new prior frames that form the updated prior information. Linear interpolation is used to generate new frames via column interpolation and row interpolation. This is represented as $f_{MC}(\cdot)$ in Lines \ref{motion compensation 1} and \ref{motion compensation 2} in the Algorithm \ref{CORPCA-OFAlg}. By using the flow fields $(\mathsf{\textbf{v}}_{1\mathsf{x}}, \mathsf{\textbf{v}}_{1\mathsf{y}})$ and $(\frac{1}{2}\mathsf{\textbf{v}}_{2\mathsf{x}}, \frac{1}{2}\mathsf{\textbf{v}}_{2\mathsf{y}})$ to predict motions in the next frame and compensate them on $\bx_{t-1}$, we obtain $\bx'_{t-2}$ and $\bx'_{t-3}$, respectively. Here $\bx'_{t-3}$ is obtained by compensating for the half of motions, i.e., $(\frac{1}{2}\mathsf{\textbf{v}}_{2\mathsf{x}}, \frac{1}{2}\mathsf{\textbf{v}}_{2\mathsf{y}})$, between $\bx_{t-1}$ and $\bx_{t-3}$. These generated frames $\bx'_{t-2}$, $\bx'_{t-3}$ are more correlated to the current frame $\bx_t$ than $\bx_{t-2}$, $\bx_{t-3}$. We also keep the most recent frame $\bx'_{t-1}=\bx_{t-1}$ (in Line \ref{keepPrior}) as one of the prior frames. 

Thereafter, $\bv_{t}^{(k+1)}$ and $\bx_{t}^{(k+1)}$ are iteratively computed as in Lines \ref{endV}-\ref{endX} in Algorithm \ref{CORPCA-OFAlg} to solve Problem \eqref{CORPCAminimization}. It can be noted that the proximal operator $\mathbf{\Gamma}_{\tau g_{1}}(\cdot)$ in Line \ref{gamma} of Algorithm 1 is defined \cite{LuongARXIV17} as
\vspace{-0.4pt}
\begin{equation}\label{l1-proximalOperatorMatrix}
\mathbf{\Gamma}_{\tau g_{1}}(\bX) = \argmin_{\bV }\Big\{ \tau g_{1}(\bV) + \frac{1}{2}||\bV-\bX||^{2}_{2}\Big\},
\vspace{-0.4pt}
\end{equation}
where $g_{1}(\cdot)\hspace{-2pt}=\hspace{-2pt}\|\cdot\|_{1}$. The weights $\mathbf{W}_{j}$ and $\beta_{j}$ are updated per iteration of the algorithm (see Lines \ref{weightW}-\ref{weightBeta}). As suggested in \cite{JWright09}, the convergence of Algorithm 1 in Line \ref{converged} is determined by evaluating the criterion 
\vspace{-0pt}
$\|\partial H(\bx_{t},\bv_{t})|_{\bx_{t}^{(k+1)},\bv_{t}^{(k+1)}}\|_{2}^{2}\hspace{-2pt}<\hspace{-2pt}2*10^{-7}\|(\bx_{t}^{(k+1)},\bv_{t}^{(k+1)})\|_{2}^{2}.\vspace{-0pt}$ After this, we update the prior information for the next instance, $\bZ_{t}$ and $\bB_{t}$, in Lines \ref{updateZ}-\ref{updateB}. 
\begin{figure*}[tp!]
	\centering
	\subfigure[\vspace{-0.15pt}\texttt{Bootstrap} \#2213]{
		\includegraphics[width=0.475\textwidth]{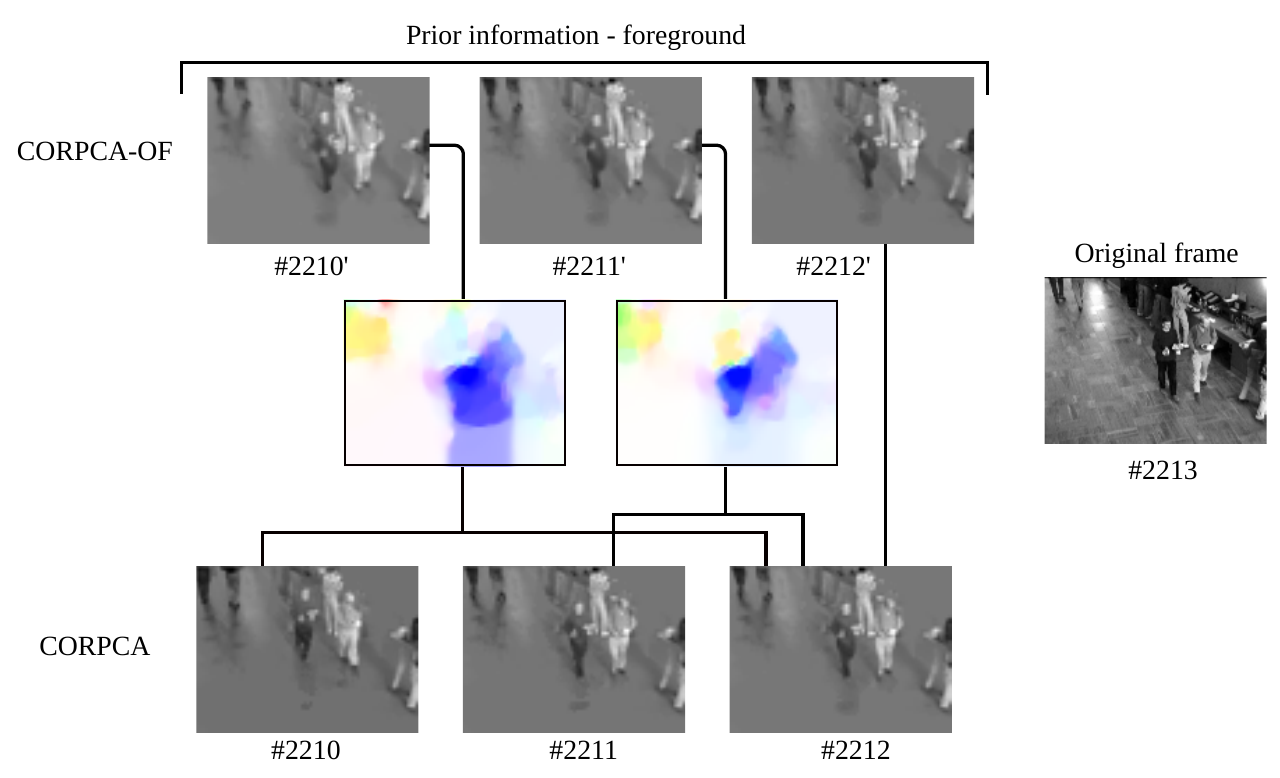}\hspace{7pt}\label{bootstrap2213}}
	%
	\subfigure[\vspace{-0.15pt}\texttt{Curtain} \#2866]{
		\includegraphics[width=0.475\textwidth]{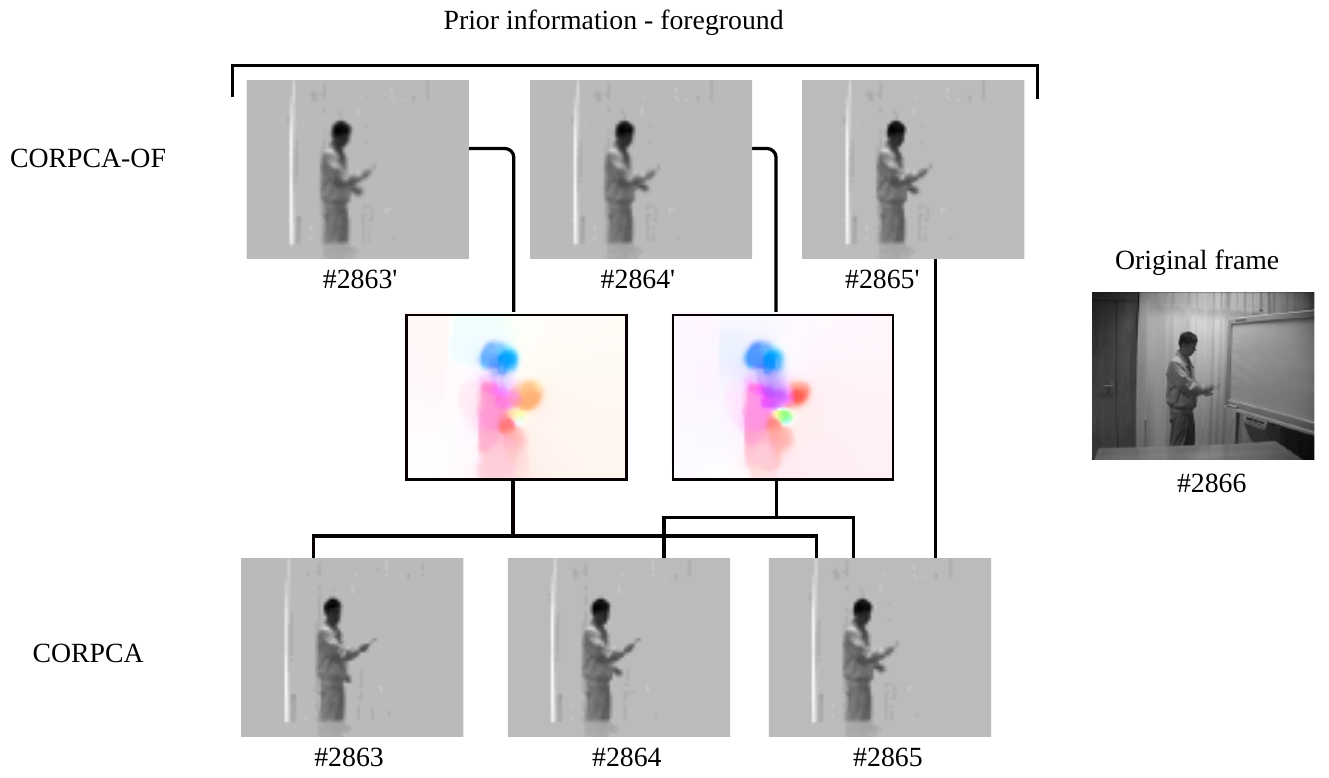}\hspace{7pt}\label{curtain2866}}
	
	\subfigure[\vspace{-0.15pt}\texttt{Bootstrap} \#451]{
		\includegraphics[width=0.475\textwidth]{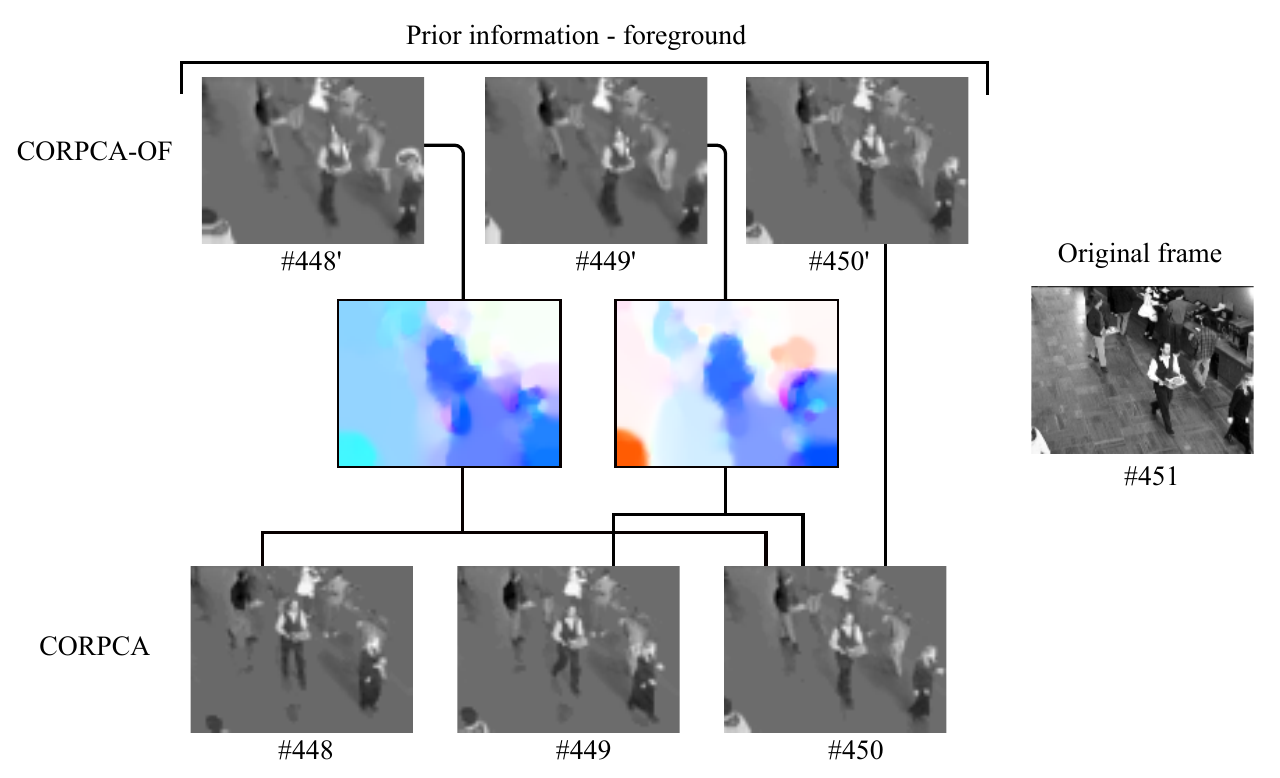}\hspace{7pt}\label{bootstrap451}}	
	\subfigure[\vspace{-0.15pt}\texttt{Curtain} \#2774]{
		\includegraphics[width=0.475\textwidth]{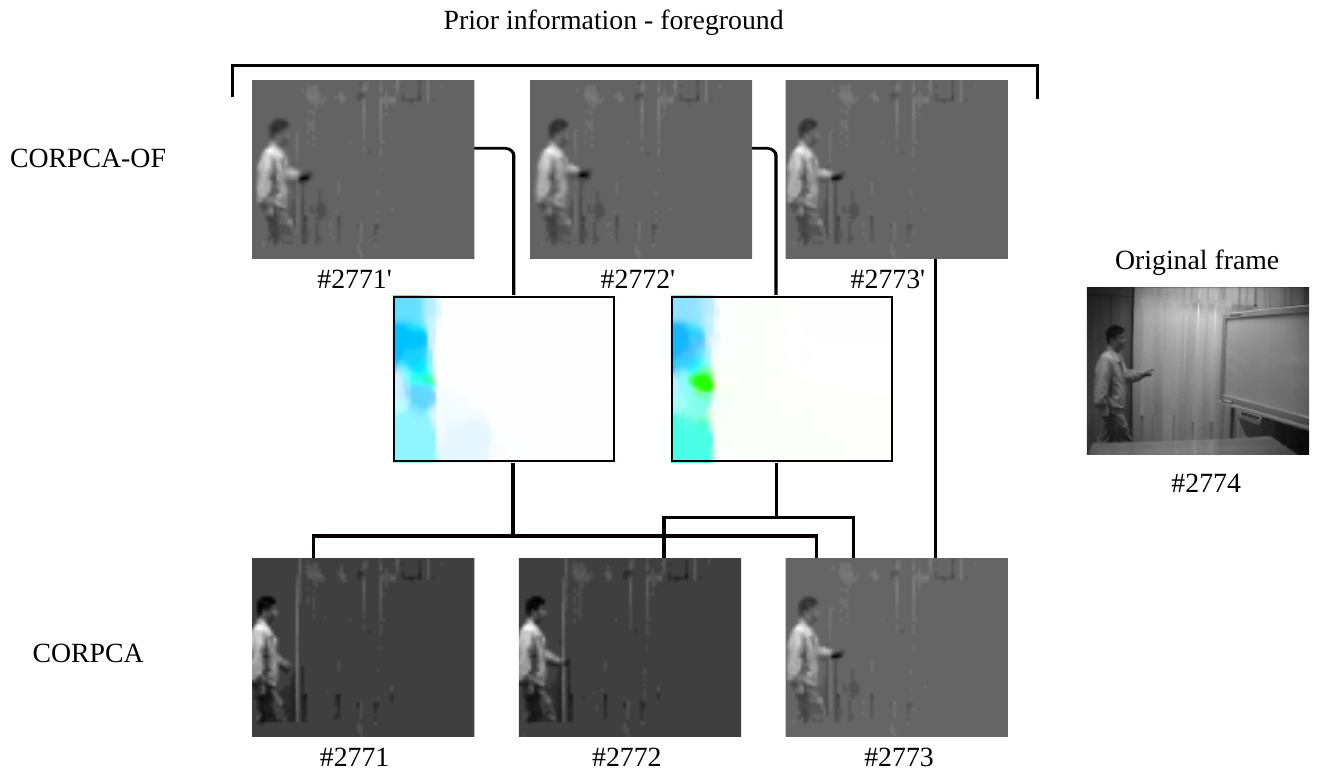}\hspace{7pt}\label{curtain2774}}
	\caption{Prior information generation in CORPCA-OF using optical flow \cite{brox2011}. 
	}\label{corpcaOF_results}
	
	\vspace{-0pt}
\end{figure*}
\setlength{\textfloatsep}{0pt}
\begin{algorithm}[tp!]
	\DontPrintSemicolon \SetAlgoLined
	\nonl\textbf{Input}: $\by_{t},~\mathbf{\Phi},~\bZ_{t-1},~\bB_{t-1}$; \\ 
	\nonl\textbf{Output}: $\bhx_{t},~\bhv_{t},~\bZ_{t},~\bB_{t}$;\\
	\tcp{Initialize variables and parameters.}
	$\bx_{t}^{(-1)}\hspace{-2pt}=\hspace{-2pt}\bx_{t}^{(0)}\hspace{-2pt}=\hspace{-2pt}\mathbf{0}$; $\bv_{t}^{(-1)}\hspace{-2pt}=\hspace{-2pt}\bv_{t}^{(0)}\hspace{-2pt}=\hspace{-2pt}\mathbf{0}$;
	$\xi_{-1}\hspace{-2pt}=\xi_{0}\hspace{-2pt}=\hspace{-2pt}1$; $\mu_{0}\hspace{-2pt}=\hspace{-2pt}0$; $\bar{\mu}\hspace{-2pt}>\hspace{-2pt}0$; $\lambda>0$; $0\hspace{-2pt}<\hspace{-2pt}\epsilon\hspace{-2pt}<\hspace{-2pt}1$; $k\hspace{-2pt}=\hspace{-2pt}0$; $g_{1}(\cdot)\hspace{-2pt}=\hspace{-2pt}\|\cdot\|_{1}$; \\
	\tcp{Motion estimation and compensation using Large Displacement Optical Flow \cite{brox2011}}
	$ (\mathsf{\textbf{v}}_{1\mathsf{x}}, \mathsf{\textbf{v}}_{1\mathsf{y}}) = f_{ME}(\bx_{t-1}, \bx_{t-2}); $ \label{motion estimation 1}\\
	$ (\mathsf{\textbf{v}}_{2\mathsf{x}}, \mathsf{\textbf{v}}_{2\mathsf{y}}) = f_{ME}(\bx_{t-1}, \bx_{t-3}); $\label{motion estimation 2}
	\\
	$ \bx'_{t-1} = \bx_{t-1}; $\label{keepPrior}
	\\
	$ \bx'_{t-2} = f_{MC}(\bx_{t-1}, \mathsf{\textbf{v}}_{1\mathsf{x}}, \mathsf{\textbf{v}}_{1\mathsf{y}}); $ \label{motion compensation 1}
	\\
	$ \bx'_{t-3} = f_{MC}(\bx_{t-1},\frac{1}{2} \mathsf{\textbf{v}}_{2\mathsf{x}}, \frac{1}{2} \mathsf{\textbf{v}}_{2\mathsf{y}}); $\label{motion compensation 2}
	\\	
	$\bz_{J}=\bx'_{t-1}; ~\bz_{J-1}=\bx'_{t-2}; ~\bz_{J-2}=\bx'_{t-3}$;\\
	\While{not converged}{\label{converged}
		\tcp{Solve Problem \eqref{CORPCAminimization}.}
		$\widetilde{\bv_{t}}^{(k)}\hspace{-2pt}=\bv_{t}^{(k)}\hspace{-2pt}+\hspace{-2pt}\frac{\xi_{k-1}-1}{\xi_{k}}(\bv_{t}^{(k)}\hspace{-2pt}-\hspace{-2pt}\bv_{t}^{(k-1)})$; \label{startV}\\
		$\widetilde{\bx_{t}}^{(k)}\hspace{-2pt}=\bx_{t}^{(k)}\hspace{-2pt}+\hspace{-2pt}\frac{\xi_{k-1}-1}{\xi_{k}}(\bx_{t}^{(k)}\hspace{-2pt}-\hspace{-2pt}\bx_{t}^{(k-1)})$; \\
		$\nabla_{\bv_{t}} f(\widetilde{\bv_{t}}^{(k)},\widetilde{\bx_{t}}^{(k)})=\nabla_{\bx_{t}}f(\widetilde{\bv_{t}}^{(k)},\widetilde{\bx_{t}}^{(k)})=\mathbf{\Phi}^{\mathrm{T}}\Big(\mathbf{\Phi} (\widetilde{\bv_{t}}^{(k)}+\widetilde{\bx_{t}}^{(k)})-\by_{t}\Big)$; \\ 
		
		%
		%

		\hspace{-0pt}$(\bU_{t},\bSi_{t},\bV_{t})\hspace{-1pt}=\hspace{-1pt}\hspace{-2pt}\mathrm{incSVD}\Big(\hspace{-0pt}\Big[\bB_{t-1}~\Big(\widetilde{\bv_{t}}^{(k)}\hspace{-2pt}-\hspace{-2pt}\frac{1}{2}\nabla_{\bv_{t}} f(\widetilde{\bv_{t}}^{(k)},\widetilde{\bx_{t}}^{(k)})\Big)\Big]\Big)$; \label{incSVD}\\
		
		$\bT_{t}\hspace{-2pt}=\hspace{-2pt}\bU_{t}\mathbf{\Gamma}_{\frac{\mu_{k}}{2}g_{1}}(\bSi_{t})\bV_{t}^{T}$;\label{gamma}\\ 
		$\bv_{t}^{(k+1)}\hspace{-2pt}=\bT_{t}(:,\mathrm{end})$;\label{endV} \\
		$\bx_{t}^{(k+1)}\hspace{-2pt}=\hspace{-2pt}\Gamma_{\frac{\mu_{k}}{2}g}\Big(\widetilde{\bx_{t}}^{(k)}-\frac{1}{2}\nabla_{\bx_{t}} f(\widetilde{\bv_{t}}^{(k)},\widetilde{\bx_{t}}^{(k)})\Big)$; where $\Gamma_{\frac{\mu_{k}}{2}g}(\cdot)$ is given as in RAMSIA \cite{LuongICIP16};\label{endX}\\
		\tcp{Compute the updated weights \cite{LuongICIP16}\hspace{-2pt}.}
		\vspace{-0.2pt}
		$w_{ji} \hspace{-2pt}=\dfrac{n(|x_{ti}^{(k+1)}\hspace{-2pt}-\hspace{-2pt}z_{ji}|\hspace{-2pt}+\hspace{-1pt}\epsilon)^{-1}}{\sum\limits_{l=1}^{n}(|x_{tl}^{(k+1)}\hspace{-2pt}-\hspace{-2pt}z_{jl}|\hspace{-2pt}+\hspace{-1pt}\epsilon)^{-1}}$;\label{weightW}
		\vspace{-0.1pt}
		\\
		$\beta_{j}\hspace{-2pt} =\dfrac{\Big(||\mathbf{W}_{j}(\bx_{t}^{(k+1)}\hspace{-2pt}-\hspace{-2pt}\bz_{j})||_{1}\hspace{-2pt}+\hspace{-1pt}\epsilon\Big)^{-1}}{\sum\limits_{l=0}^{J}\hspace{-2pt}\Big(||\mathbf{W}_{l}(\bx_{t}^{(k+1)}\hspace{-2pt}-\hspace{-2pt}\bz_{l})||_{1}\hspace{-2pt}+\hspace{-1pt}\epsilon\Big)^{-1}}$;\label{weightBeta}
		
		$\xi_{k+1}=(1+\sqrt{1+4\xi_{k}^{2}})/2$; $\mu_{k+1}=\max(\epsilon\mu_{k},\bar{\mu})$;\\
		$k=k+1$; \\
	}
	\vspace{-0.3pt}
	\tcp{Update prior information.}
	$\bZ_{t}:=\{\bz_{j}=\bx^{(k+1)}_{t-J+j}\}_{j=1}^{J}$;\label{updateZ}\\
	$\bB_{t}=\bU_{t}(:,1:d)\mathbf{\Gamma}_{\frac{\mu_{k}}{2}g_{1}}(\bSi_{t})(1:d,1:d)\bV_{t}(:,1:d)^{\mathrm{T}}$;\label{updateB}\\
	
	\Return $\bhx_{t}=\bx_{t}^{(k+1)},~\bhv_{t}=\bv_{t}^{(k+1)},~\bZ_{t},~\bB_{t}$;
	\caption{The proposed CORPCA-OF algorithm.}
	\label{CORPCA-OFAlg}
\end{algorithm}
\vspace{-0pt}

\textbf{Prior Update}. The update of $\bZ_{t}$ and $\bB_{t}$ \cite{LuongARXIV17} is carried out after each time instance (see Lines \ref{updateZ}-\ref{updateB}, Algorithm 1). Due to the correlation between subsequent frames, we update the prior information $\bZ_{t}$ by using the $J$ latest recovered sparse components, which is given by, $\bZ_{t}:=\{\bz_{j}=\bx_{t-J+j}\}_{j=1}^{J}$. For $\bB_{t}\in \mathbb{R}^{n\times d}$, we consider an adaptive update, which operates on a fixed or constant number $d$ of the columns of $\bB_{t}$. To this end, the incremental singular decomposition $\mathrm{SVD}$ \cite{Brand02} method ($\mathrm{incSVD}(\cdot)$ in Line \ref{incSVD}, Algorithm \ref{CORPCA-OFAlg}) is used. It is worth noting that the update $\bB_{t}=\bU_{t}\mathbf{\Gamma}_{\frac{\mu_{k}}{2}g_{1}}(\bSi_{t})\bV_{t}^{\mathrm{T}}$, causes the dimension of $\bB_{t}$ to increase as $\bB_{t}\in \mathbb{R}^{n\times (d+1)}$ after each instance. However, in order to maintain a reasonable number of $d$, we take $\bB_{t}=\bU_{t}(:,1:d)\mathbf{\Gamma}_{\frac{\mu_{k}}{2}g_{1}}(\bSi_{t})(1:d,1:d)\bV_{t}(:,1:d)^{\mathrm{T}}$. The computational cost of $\mathrm{incSVD}(\cdot)$ is lower than conventional SVD \cite{Brand02,Rodriguez16} since we only compute the full $\mathrm{SVD}$ of the middle matrix with size $(d+1)\times(d+1)$, where $d\ll n$, instead of $n\times(d+1)$.

The computation of $\mathrm{incSVD}(\cdot)$ is presented in the following: The goal is to compute $\mathrm{incSVD}[\bB_{t-1} ~\bv_{t}]$, i.e., $[\bB_{t-1} ~\bv_{t}]=\bU_{t}\bSi_{t}\bV_{t}^\mathrm{T}$. By taking the SVD of $\bB_{t-1}\in \mathbb{R}^{n\times d}$ to obtain $\bB_{t-1}=\bU_{t-1}\bSi_{t-1}\bV_{t-1}^\mathrm{T}$. Therefore, we can derive $(\bU_{t},\bSi_{t},\bV_{t})$ via $(\bU_{t-1},\bSi_{t-1},\bV_{t-1})$ and $\bv_{t}$. We write the matrix $[\bB_{t-1} ~\bv_{t}]$ as
\begin{equation}\label{incSVDUpdate}
\hspace{-1pt}[\bB_{t-1} ~\bv_{t}]\hspace{-2pt}=\hspace{-2pt}\Big[\bU_{t-1}~\dfrac{\bdel_{t}}{\|\bdel_{t}\|_{2}}\Big]\hspace{-2pt}\cdot\hspace{-2pt}\left[\hspace{-2pt}\begin{array}{cc} \bSi_{t-1}\hspace{-2pt} &\hspace{-2pt} \be_{t}\\ \textbf{0}^\mathrm{T}\hspace{-2pt} & \hspace{-2pt}\|\bdel_{t}\|_{2} \end{array}\hspace{-2pt}\right]
\hspace{-2pt}\cdot\hspace{-2pt}\left[\hspace{-2pt}\begin{array}{cc} \bV_{t-1}^\mathrm{T} \hspace{-2pt}&\hspace{-2pt} \textbf{0}\\ \textbf{0}^\mathrm{T} \hspace{-2pt}&\hspace{-2pt} 1 \end{array}\hspace{-2pt}\right],
\end{equation}
where $\be_{t}=\bU_{t-1}^\mathrm{T}\bv_{t}$ and $\bdel_{t} =\bv_{t}-\bU_{t-1}\be_{t}$.
By taking the $\mathrm{SVD}$ of the matrix in between the right side of \eqref{incSVDUpdate}, we obtain $\left[\begin{array}{cc} \bSi_{t-1}\hspace{-2pt} &\hspace{-2pt} \be_{t}\\ \textbf{0}^\mathrm{T} \hspace{-2pt}& \hspace{-2pt}\|\bdel_{t}\|_{2} \end{array}\right]= \widetilde{\bU}\widetilde{\bSi}\widetilde{\bV}^\mathrm{T}$.
Eventually, we obtain $\bU_{t}=\Big[\bU_{t-1}~\dfrac{\bdel_{t}}{\|\bdel_{t}\|_{2}}\Big]\cdot\widetilde{\bU}\label{incSVDFinalU}$, $\bSi_{t}\hspace{-2pt}=\hspace{-2pt}\widetilde{\bSi}$, and $\bV_{t}=\left[\begin{array}{cc} \bV_{t-1}^\mathrm{T}\hspace{-2pt} &\hspace{-2pt} \textbf{0}\\ \textbf{0}^\mathrm{T}\hspace{-2pt} &\hspace{-2pt} 1 \end{array}\right]\cdot\widetilde{\bV}$.

\hspace{4pt}

\section{Experimental Results}
\label{Experiment}

\hspace{4pt}
%

We evaluate the performance of our proposed CORPCA-OF in Algorithm \ref{CORPCA-OFAlg} and compare CORPCA-OF against the existing methods, RPCA \cite{CandesRPCA}, GRASTA \cite{JHe12}, and ReProCS \cite{GuoQV14}. RPCA \cite{CandesRPCA} is a batch-based method assuming full access to the data, while GRASTA \cite{JHe12} and ReProCS \cite{GuoQV14} are online methods that can recover either the (low-rank) background component (GRASTA) or the (sparse) foreground component (ReProCS) from compressive measurements. In this work, we test two sequences \cite{Li04}, \texttt{Bootstrap} (60$\times$80 pixels) and \texttt{Curtain} (64$\times$80 pixels), having a static and a dynamic background, respectively.
\vspace{-0.2pt}


%
\vspace{-1pt}
\subsection{Prior Information Evaluation}
We evaluate the prior information of CORPCA-OF compared with that of CORPCA\cite{LuongARXIV17} using the previously separated foreground frames directly. For CORPCA-OF, we generate the prior information by estimating and compensating motions among the previous foreground frames. Fig. \ref{corpcaOF_results} shows a few examples of the prior information generated for the sequences \texttt{Bootstrap} and \texttt{Curtain}. In Fig. \ref{bootstrap2213}, it can be observed that frames \#2210', \#2211' and \#2212' (of CORPCA-OF) are better than corresponding \#2210, \#2211 and \#2212 (of CORPCA) for the current frame \#2213, similarly in Figs. \ref{curtain2866}, \ref{bootstrap451}, and \ref{curtain2774}. Specially, in Fig. \ref{bootstrap451}, the generated frames \#448' and \#449' are significantly improved due to compensating the given dense motions. In Fig. \ref{curtain2774}, it is clear that the movements of the person is well compensated in \#2771' and \#2772' by CORPCA-OF compared to \#2771 and \#2772 respectively, of CORPCA, leading to better correlations with the foreground of current frame \#2774.

\vspace{-0.2pt}
\subsection{Compressive Video Foreground and Background Separation}
\begin{figure}[tp!]
	\centering
	\subfigure[\vspace{-0.15pt}\texttt{Bootstrap}]{
		\includegraphics[width=0.48\textwidth]{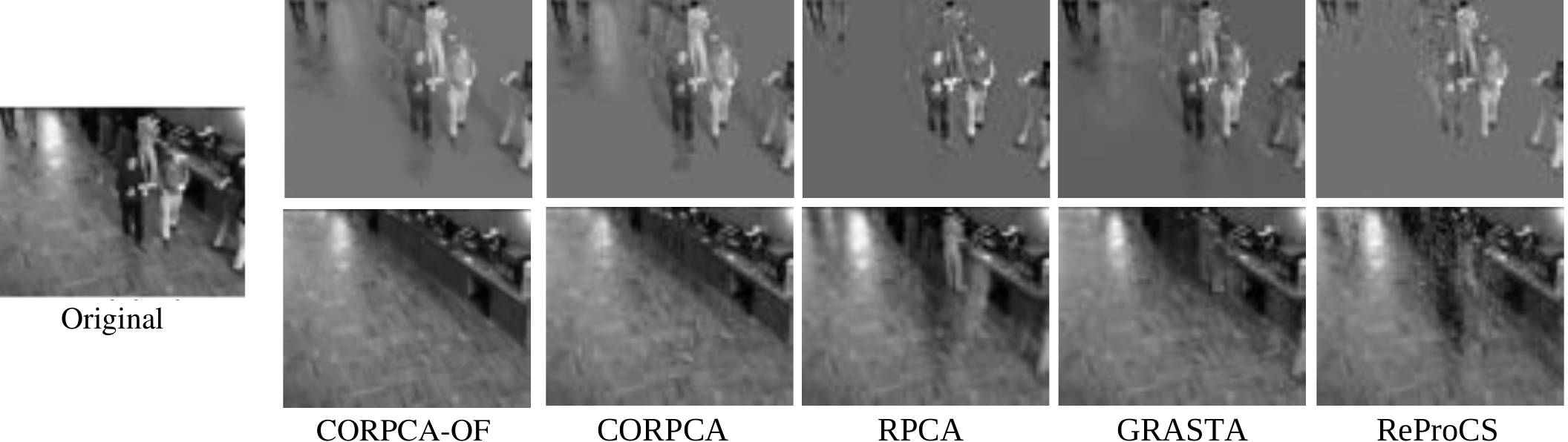}\hspace{1.5pt}\label{bootstrapRateFull}}
	%
	\subfigure[\vspace{-0.15pt}\texttt{Curtain}]{
		\includegraphics[width=0.48\textwidth]{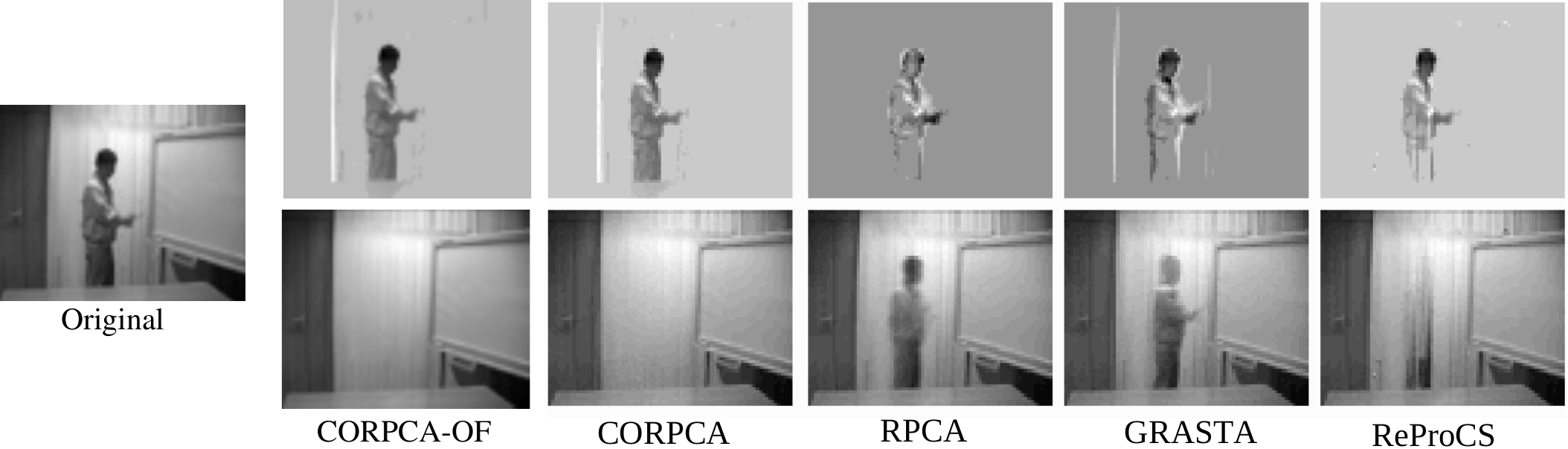}\hspace{1.5pt}\label{curtainRateFull}}
	\vspace{-0.4pt}
	\caption{Foreground and background separation for the different separation methods with full data access \texttt{Bootstrap} \#2213 and \texttt{Curtain} \#2866.
	}\label{figVisualPerform}
	
	\vspace{-0pt}
\end{figure}
\begin{figure}[t!]
	\centering
	\subfigure[\vspace{-0.15pt}CORPCA-OF]{
		\includegraphics[width=0.42\textwidth]{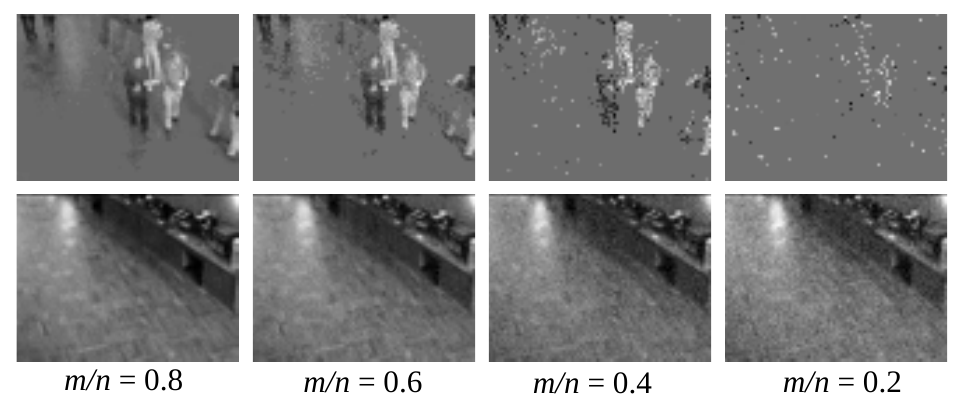}\label{CORPCA-OF_bootstrap}}
	\subfigure[\vspace{-0.15pt}CORPCA \cite{LuongARXIV17}]{
		\includegraphics[width=0.42\textwidth]{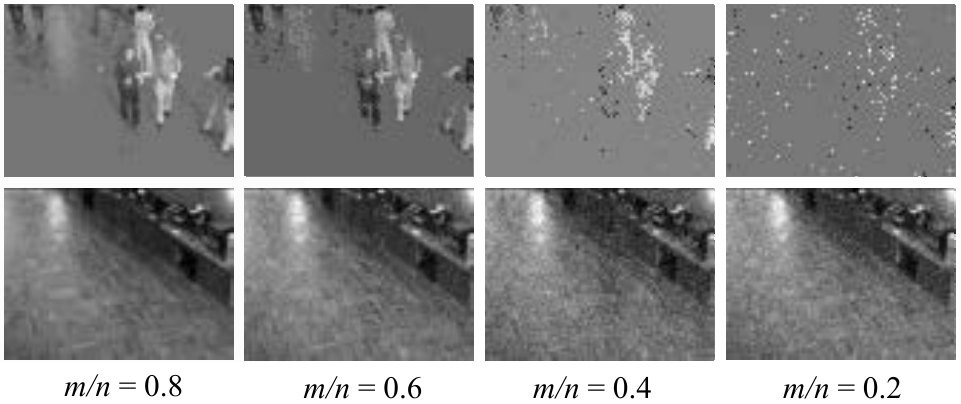}\label{CORPCA_bootstrap}}
	\subfigure[\vspace{-0.15pt}ReProCS \cite{GuoQV14}]{
		\includegraphics[width=0.42\textwidth]{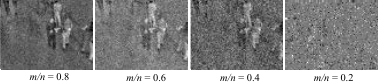}\label{ReProCS_bootstrap}}
	\caption{Compressive foreground and background separation of CORPCA-OF, CORPCA \cite{LuongARXIV17}, and ReProCS \cite{GuoQV14} with different measurement rates $m/n$ of frame \texttt{Bootstrap} \#2213.
	}\label{figVisualCompressedBootstrap}
	\vspace{-0pt}
\end{figure}
\begin{figure}[tp!]
	\centering
	\subfigure[\vspace{-0.15pt}CORPCA-OF]{
		\includegraphics[width=0.4\textwidth]{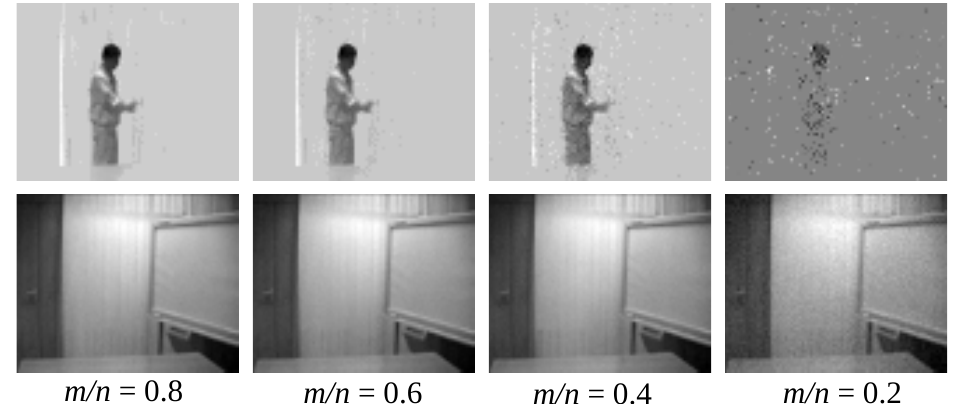}\label{CORPCA-OF_curtain}}
	\subfigure[\vspace{-0.15pt}CORPCA \cite{LuongARXIV17}]{
		\includegraphics[width=0.4\textwidth]{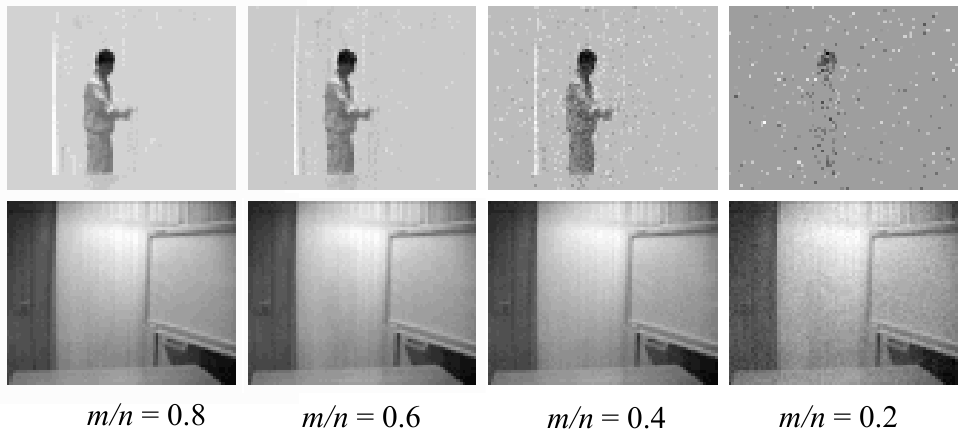}\label{CORPCA_curtain}}
	\subfigure[\vspace{-0.15pt}ReProCS \cite{GuoQV14}]{
		\includegraphics[width=0.4\textwidth]{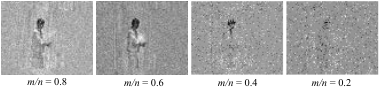}\label{ReProCS_curtain}}
	\caption{Compressive foreground and background separation of CORPCA-OF, CORPCA \cite{LuongARXIV17}, and ReProCS \cite{GuoQV14} with different measurement rates $m/n$ of frame \texttt{Curtain} \#2866.
	}\label{figVisualCompressedCurtain}
\end{figure}
\vspace{-0.4pt}
\begin{figure*}[tp!]
	\centering
	\subfigure[\vspace{-0.15pt}\texttt{Bootstrap}]{
		\includegraphics[width=0.32\textwidth]{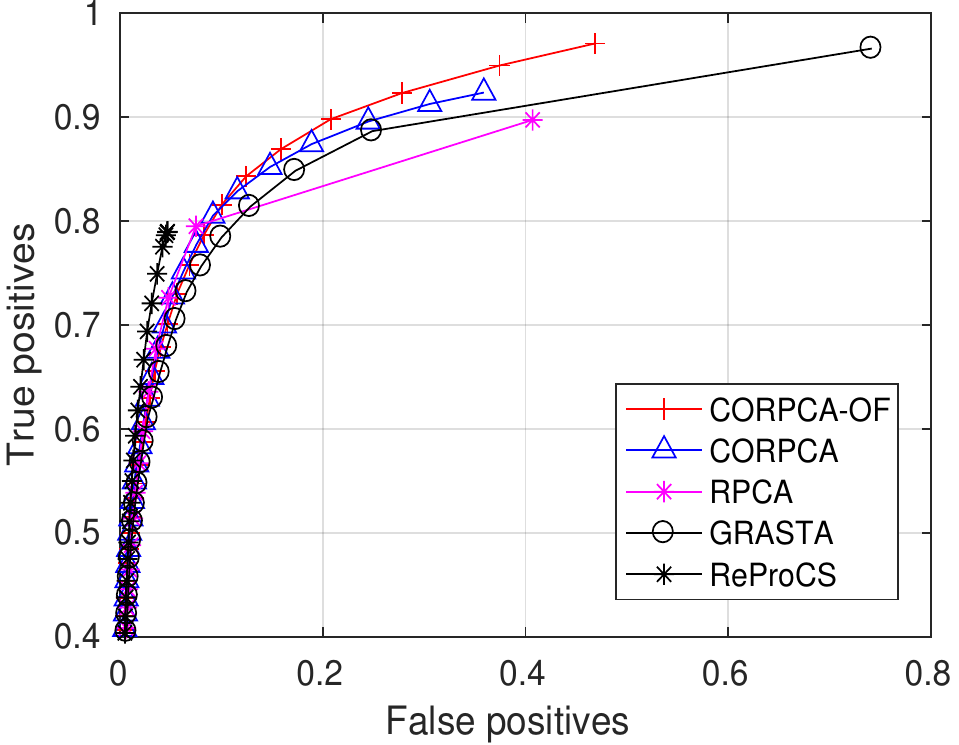}\hspace{5pt}\label{bootstrapROCFull}}
	\subfigure[\vspace{-0.15pt}\texttt{Curtain}]{
		\includegraphics[width=0.32\textwidth]{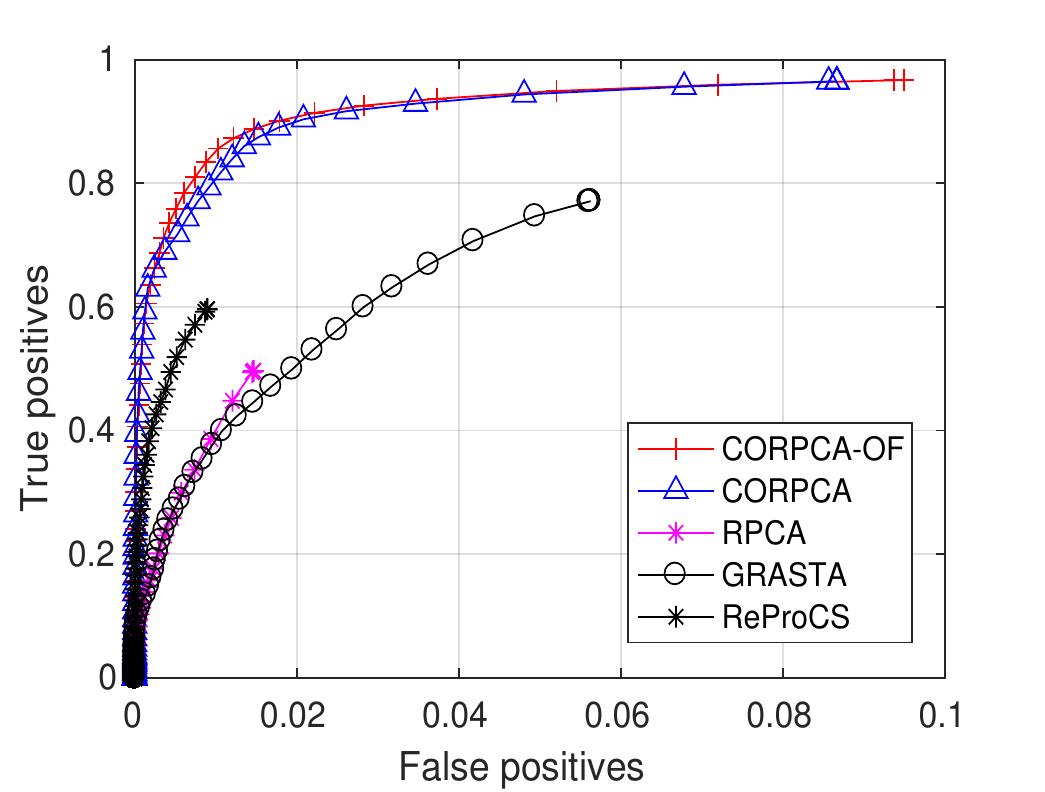}\hspace{5pt}\label{curtainROCFull}}
	\vspace{-0.4pt}
	\caption{ROC for the different separation methods with full data
		.}\label{ROCFull}
	\vspace{-0.5pt}
\end{figure*}
\vspace{-0pt}
\begin{figure*}[tp!]
	\centering
	\subfigure[\vspace{-0.15pt}CORPCA-OF]{
		\includegraphics[width=0.31\textwidth]{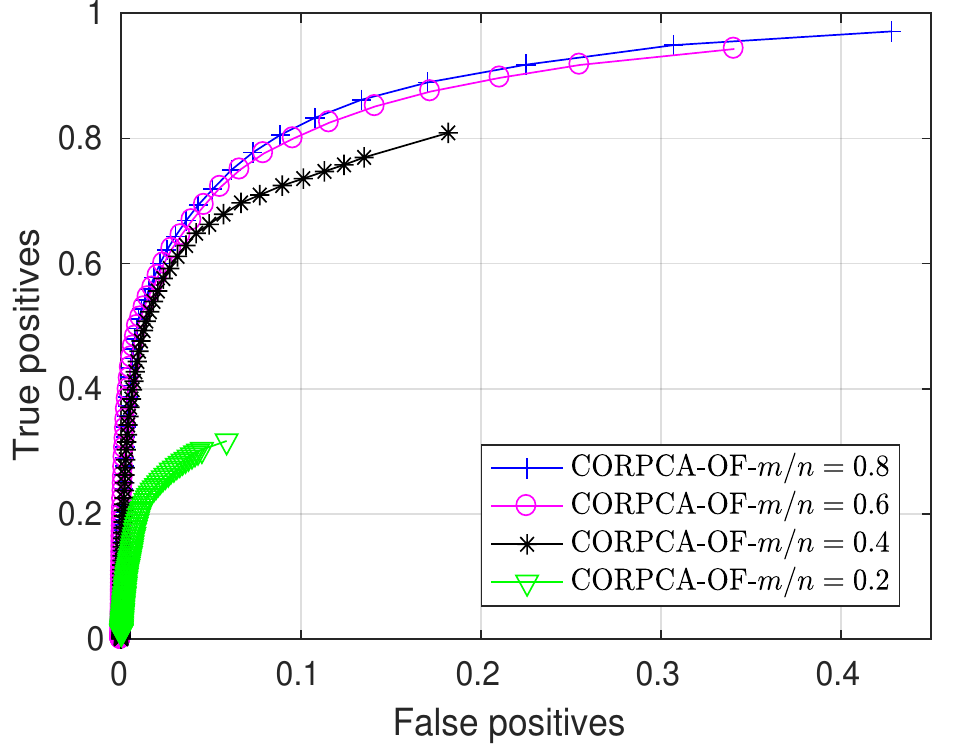}\hspace{5pt}\label{bootstrapROCCORPCA-OF}}
	\subfigure[\vspace{-0.15pt}CORPCA \cite{LuongARXIV17}]{
		\includegraphics[width=0.31\textwidth]{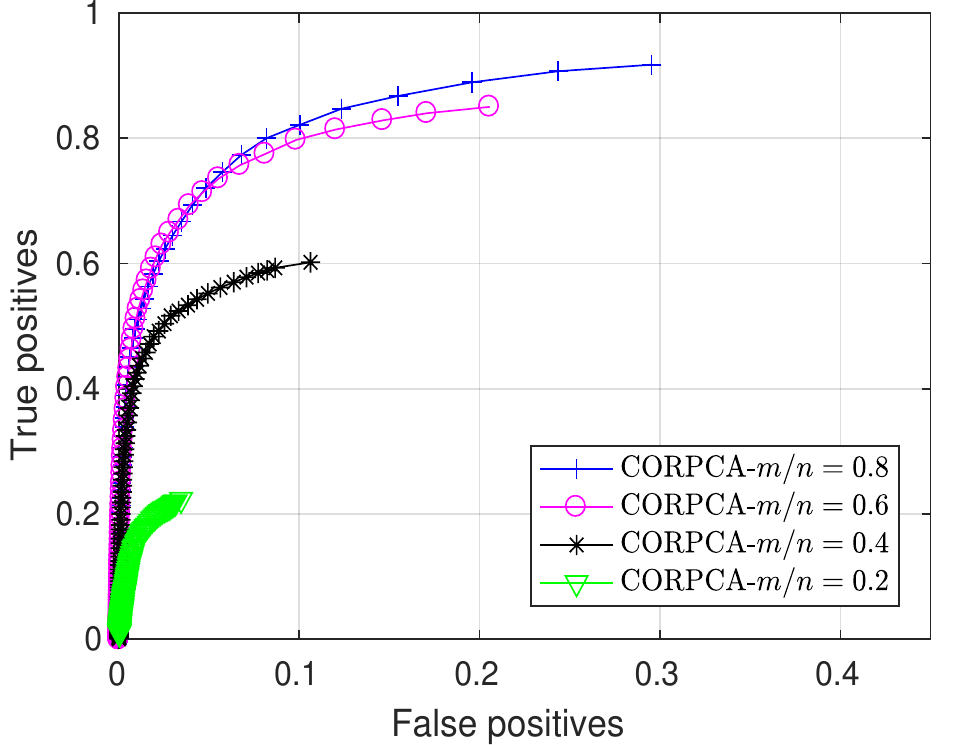}\hspace{5pt}\label{bootstrapROCCORPCA}}
	\subfigure[\vspace{-0.15pt}ReProCS \cite{GuoQV14}]{
		\includegraphics[width=0.31\textwidth]{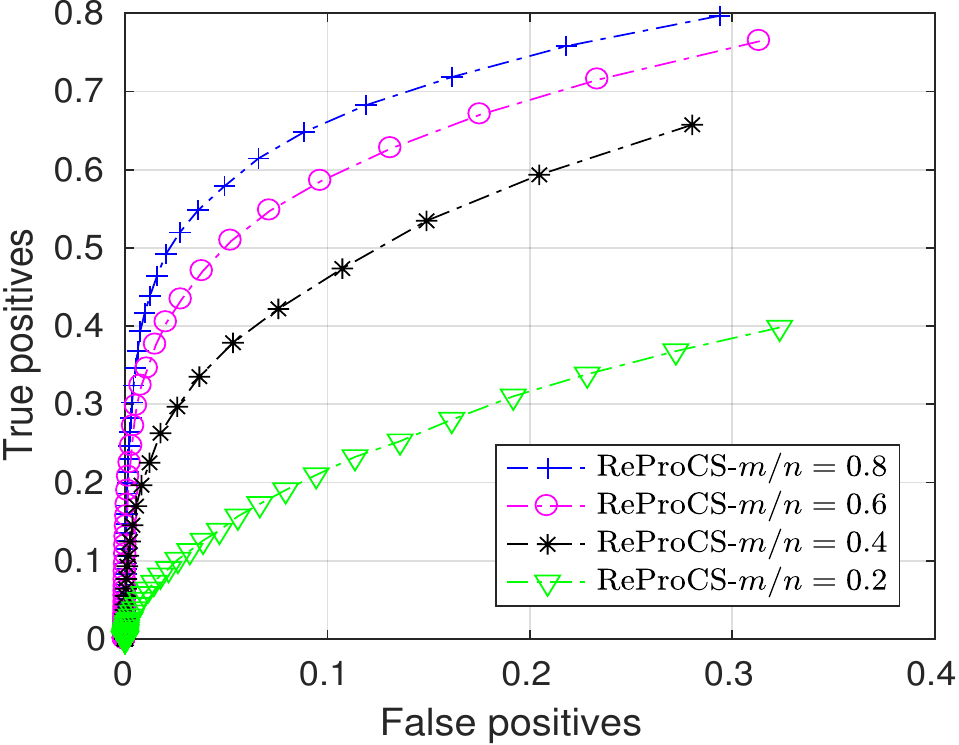}\hspace{5pt}\label{bootstrapROCReprocs}}
	\vspace{-0.4pt}
	\caption{ROC for CORPCA-OF, CORPCA \cite{LuongARXIV17}, and ReProCS \cite{GuoQV14} with compressive measurement rates $m/n$ for \texttt{Bootstrap}.}\label{ROCBootstrap}
	\vspace{-0.2pt}
\end{figure*}
\begin{figure*}[tp!]
	\centering
	
	\subfigure[\vspace{-0.15pt}CORPCA-OF]{
		\includegraphics[width=0.31\textwidth]{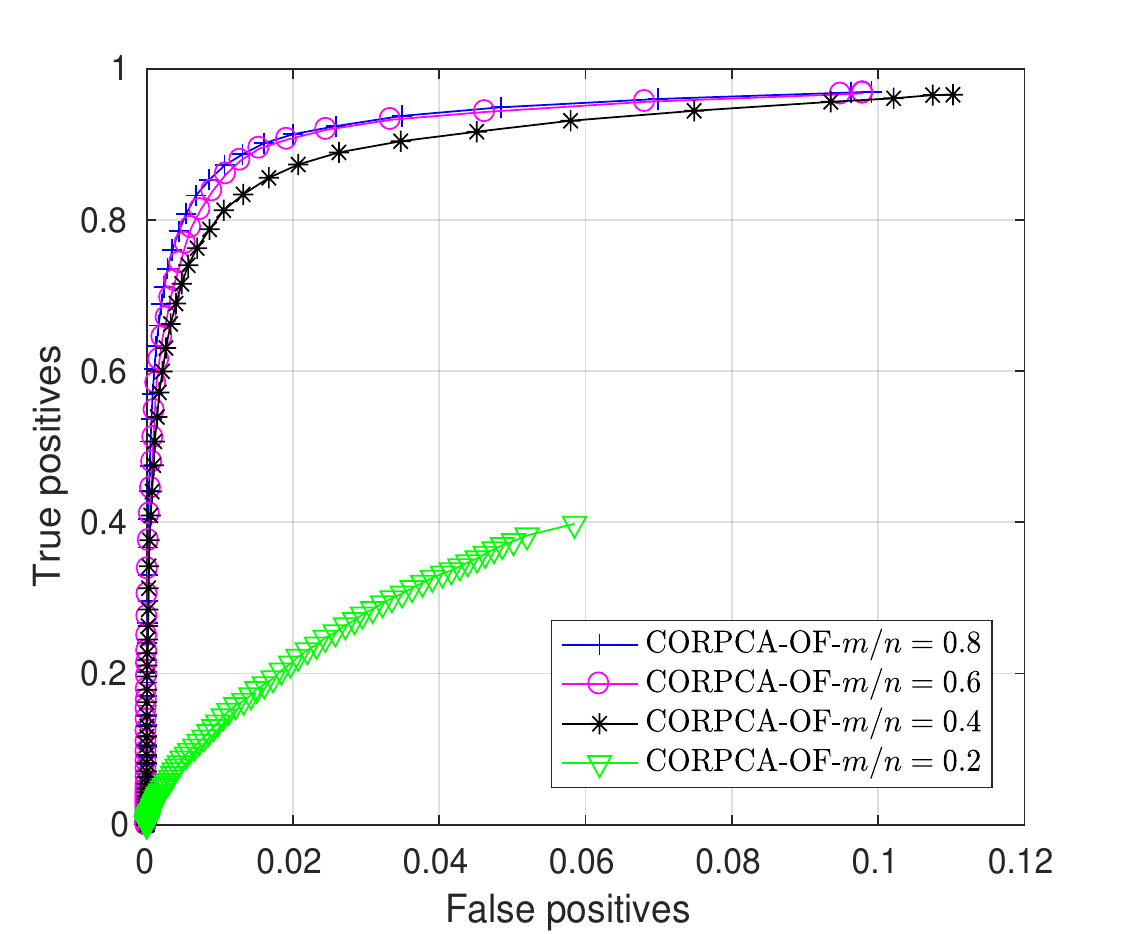}\hspace{5pt}\label{curtainROCCORPCA-OF}}
	\subfigure[\vspace{-0.15pt}CORPCA \cite{LuongARXIV17}]{
		\includegraphics[width=0.31\textwidth]{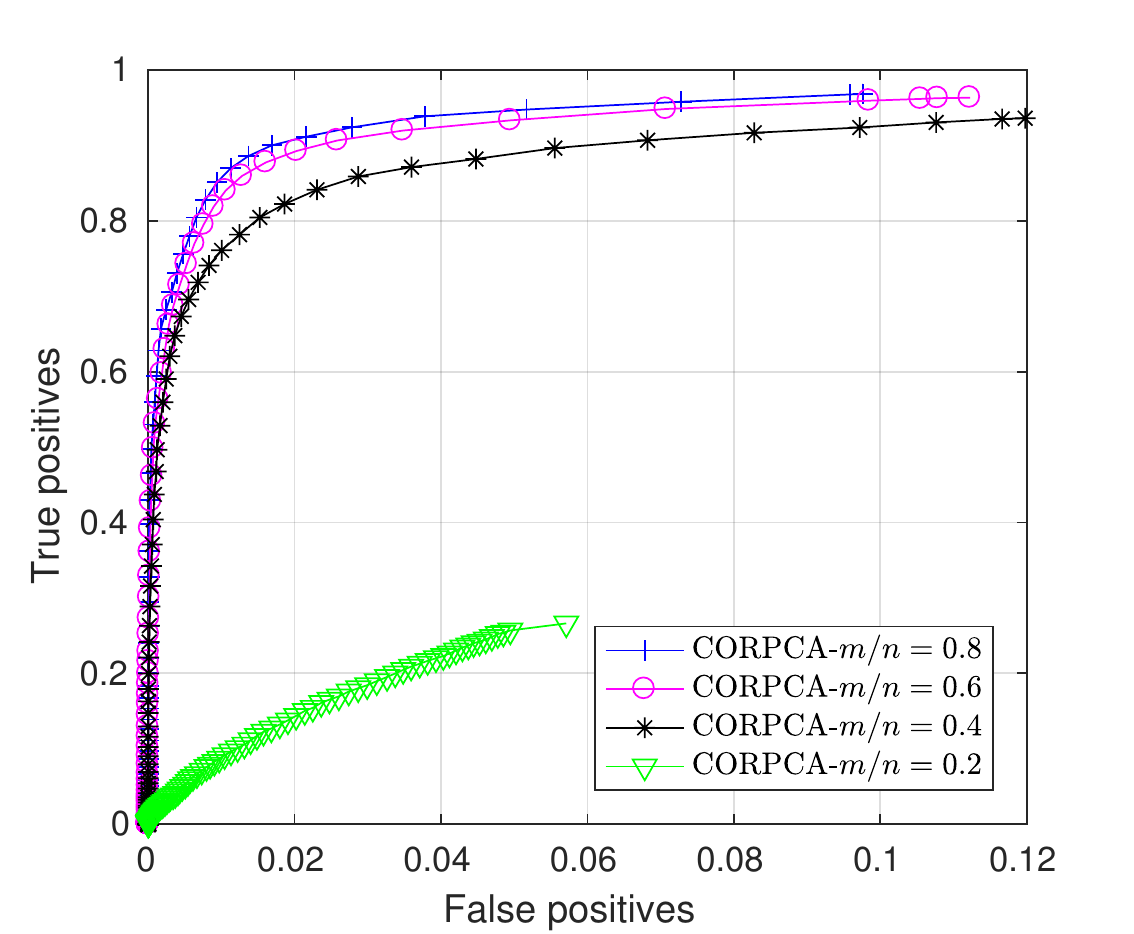}\hspace{5pt}\label{curtainROCCORPCA}}
	\subfigure[\vspace{-0.15pt}ReProCS \cite{GuoQV14}]{
		\includegraphics[width=0.31\textwidth]{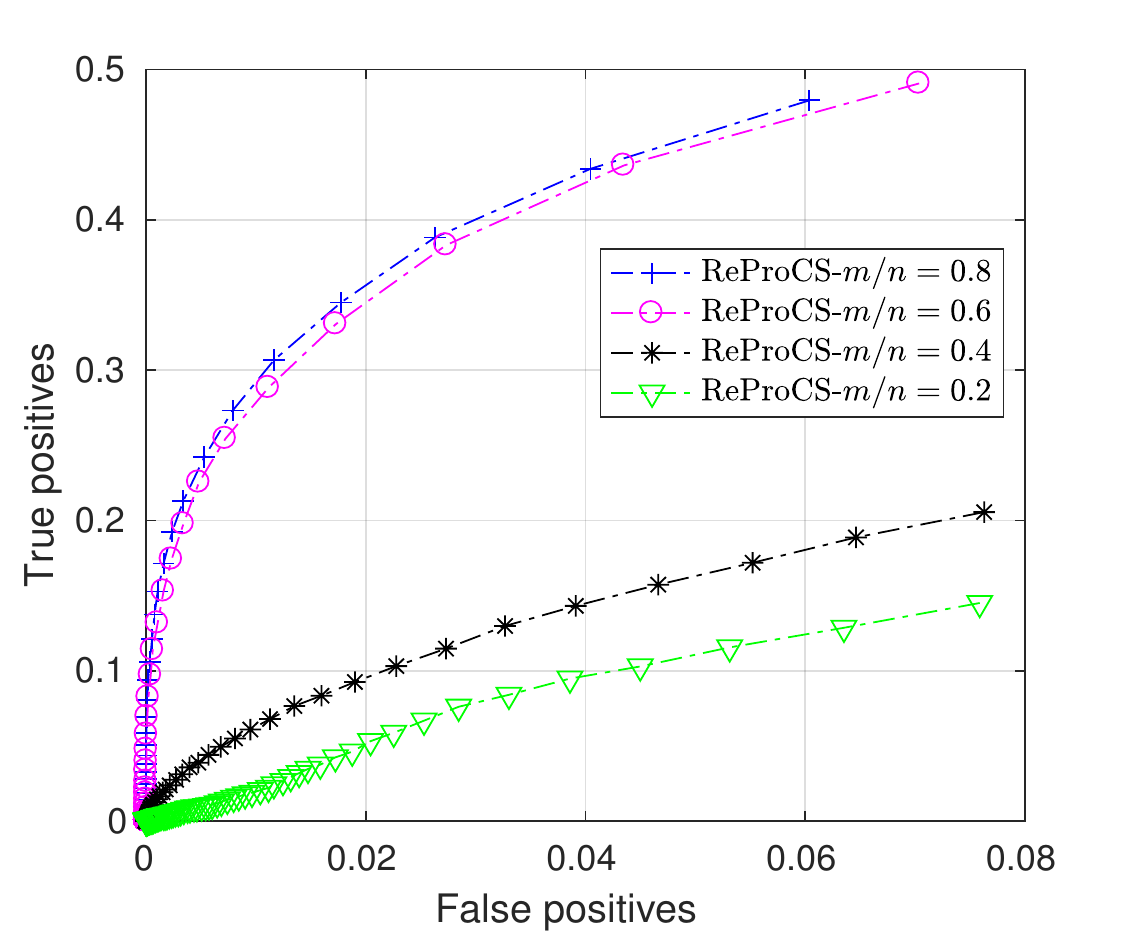}\hspace{5pt}\label{curtainROCReprocs}}
	\vspace{-0.4pt}
	\caption{ROC for CORPCA-OF, CORPCA \cite{LuongARXIV17}, and ReProCS \cite{GuoQV14} with compressive measurement rates $m/n$ for \texttt{Curtain}.}\label{ROCCurtain}
	\vspace{-0.2pt}
\end{figure*}
We assess our CORPCA-OF method in the application of compressive video separation and compare it against the existing methods, CORPCA\cite{LuongARXIV17}, RPCA \cite{CandesRPCA}, GRASTA \cite{JHe12}, and ReProCS \cite{GuoQV14}. We run all methods on the test video sequences. 
In this experiment, we use $d=100$ frames as training vectors for the proposed CORPCA-OF, CORPCA\cite{LuongARXIV17} as well as for GRASTA \cite{JHe12} and ReProCS \cite{GuoQV14}. Three latest previous foregrounds are used as the foreground prior for CORPCA, meanwhile COPRCA-OF uses them to refine the foreground prior by using optical flow \cite{brox2011}.
\vspace{-0pt}

\subsubsection{\textbf{Visual Evaluation}}

We first consider background and foreground separation with full access to the video data; the visual results of the various methods are illustrated in Fig. \ref{figVisualPerform}. It is evident that, for both the video sequences, CORPCA-OF delivers superior visual results than the other methods, which suffer from less-details in the foreground and noisy background images. We can also observe improvements over CORPCA.

Additionally, we also compare the visual results of CORPCA-OF, CORPCA and ReProCS for the frames \texttt{Bootstrap} \#2213 (in Fig. \ref{figVisualCompressedBootstrap}) and for \texttt{Curtain} \#2866 (in Fig. \ref{figVisualCompressedCurtain}) with compressed rates. They present the results under various rates on the number of measurements $m$ over the dimension $n$ of the data (the size of the vectorized frame) with rates: $m/n=\{0.8;0.6;0.4;0.2\}$. Comparing CORPCA-OF with CORPCA, we can observe in Figs. \ref{figVisualCompressedBootstrap} and \ref{figVisualCompressedCurtain} that CORPCA-OF gives the foregrounds that are less noisy and the background frames of higher visual quality. On comparison with ReProCS, our algorithm outperforms it significantly. At low rates, for instance with $m/n=0.6$ (in Fig. \ref{CORPCA-OF_bootstrap}) or $m/n=0.4$ (in Fig. \ref{CORPCA-OF_curtain}), the extracted foreground frames of CORPCA-OF are better than those of CORPCA and ReProCS. Even at a high rate of $m/n=0.8$ the sparse components or the foreground frames using ReProCS are noisy and of poor visual quality. The \texttt{Bootstrap} sequence requires more measurements than \texttt{Curtain} due to the more complex foreground information. It is evident from Figs. \ref{figVisualCompressedBootstrap} and \ref{figVisualCompressedCurtain} that the visual results obtained with CORPCA-OF are of superior quality compared to ReProCS and have significant improvements over CORPCA.

\subsubsection{\textbf{Quantitative Results}}
We evaluate quantitatively the separation performance via the \textit{receiver operating curve} (ROC) metric \cite{MDikmen}. The metrics \textit{True positives} and \textit{False positives} are defined as in \cite{MDikmen}. Fig. \ref{ROCFull} illustrates the ROC results when assuming full data access, i.e., $m/n=1$, of CORPCA-OF, CORPCA, RPCA, GRASTA, and ReProCS. The results show that CORPCA-OF delivers higher performance than the other methods.

Furthermore, we compare the foreground recovery performance of CORPCA-OF against CORPCA and ReProCS for different compressive measurement rates: $m/n=\{0.8;0.6;0.4;0.2\}$. The ROC results in Figs. \ref{ROCBootstrap} and \ref{ROCCurtain} show that CORPCA-OF can achieve higher performance in comparison to ReProCS and CORPCA. In particular, with a small number of measurements, CORPCA-OF produces better curves than those of COPRCA, e.g., for \texttt{Bootstrap} at $m/n=\{0.2;0.4;0.6\}$ [see Fig. \ref{bootstrapROCCORPCA-OF}] and for \texttt{Curtain} at $m/n=\{0.2;0.4\}$ [see Fig. \ref{curtainROCCORPCA-OF}]. The ROC results for ReProCS are quickly degraded even with a high compressive measurement rate $m/n=0.8$ [see Figure \ref{curtainROCReprocs}].

\vspace{-0pt}

\section{Conclusion}
\label{conclusion}
This paper proposed a compressive online robust PCA algorithm with optical flow (CORPCA-OF) that can process one frame per time instance using compressive measurements. CORPCA-OF efficiently incorporates multiple prior frames based on the $n$-$\ell_{1}$ minimization problem. The proposed method exploits motion estimation and compensation using optical flow to refine the prior information and obtain better quality. We have tested our method on compressive online video separation application using video data. The visual and quantitative results showed the improvements on the prior generation and the superior performance offered by CORPCA-OF compared to the existing methods including the CORPCA baseline.

\bibliographystyle{IEEEtran}
\bibliography{./IEEEfull,./IEEEabrv,./bibliography}

\end{document}